\documentclass{article}

\usepackage[preprint,nonatbib]{neurips_2023}

\usepackage{bbm}
\usepackage{xfrac}
\usepackage[utf8]{inputenc} 
\usepackage[T1]{fontenc}    
\usepackage{hyperref}       
\hypersetup{colorlinks,allcolors=black}
\usepackage{url}            
\usepackage{booktabs}       
\usepackage{nicefrac}       
\usepackage{microtype}      
\usepackage{subfiles}
\usepackage{xcolor}
\usepackage{multirow}
\usepackage{enumerate}
\usepackage{graphicx}
\usepackage{sidecap}
\usepackage[caption=false]{subfig}
\usepackage{amssymb, amsfonts, amsmath, dsfont}
\usepackage{makecell}
\usepackage{floatrow}
\usepackage{tabularx}
\usepackage[section]{placeins}
\usepackage[]{algorithm2e}

\sidecaptionvpos{figure}{t}

\DeclareRobustCommand{\Fig}[1]{Figure~\ref{fig:#1}}

\DeclareRobustCommand{\Tab}[1]{Table~\ref{tab:#1}}
\DeclareRobustCommand{\Tabs}[2]{Tables~\ref{tab:#1} and~\ref{tab:#2}}
\DeclareRobustCommand{\Eq}[1]{Equation~\ref{eq:#1}}

\def\be{\begin{equation}} 
\def\ee{\end{equation}} 
\newcommand \bea {\begin{eqnarray}} 
\newcommand \eea {\end{eqnarray}}

\newcommand{\tabitem}{~~\llap{\textbullet}~~}
\newcommand{\tabitemii}{~~\llap{$\circ$}~~}



\title{Digital Twin Generators for Disease Modeling}

\author{
    Unlearn.AI
}

\begin{document}
\maketitle

\date{\today}

\begin{abstract}
A patient's digital twin is a computational model that describes the evolution of their health over time. Digital twins have the potential to revolutionize medicine by enabling individual-level computer simulations of human health, which can be used to conduct more efficient clinical trials or to recommend personalized treatment options. Due to the overwhelming complexity of human biology, machine learning approaches that leverage large datasets of historical patients' longitudinal health records to generate patients' digital twins are more tractable than potential mechanistic models. In this manuscript, we describe a neural network architecture that can learn conditional generative models of clinical trajectories, which we call Digital Twin Generators (DTGs), that can create digital twins of individual patients. We show that the same neural network architecture can be trained to generate accurate digital twins for patients across 13 different indications simply by changing the training set and tuning hyperparameters. By introducing a general purpose architecture, we aim to unlock the ability to scale machine learning approaches to larger datasets and across more indications so that a digital twin could be created for any patient in the world.

\end{abstract} 

\section{Introduction}
\label{sec:intro}

In the near future, most important problems in medicine will be solved with computational approaches. As computational methods such as artificial intelligence (AI) improve at an astounding pace, so does our ability to model human health. Computational models of individual patients offer the opportunity to perform \emph{in silico} experiments, avoiding unnecessary trial-and-error that places a substantial burden on patients and slows down the pace of medical innovation. Digital twins are a type of specialized computational model commonly used in engineering that is increasingly being applied to healthcare. In this context, a digital twin is a computational model of a specific individual's health that is comprehensive, describing as complete a set of characteristics as possible, and which allows one to forecast that individual's future health under various scenarios of interest.

The comprehensive nature of digital twins means that they are incredibly powerful tools for solving problems in medicine that typically require experimentation on humans.  For example, when deciding which treatment a patient should receive, a patient's digital twin might allow a clinician to model, for that specific individual, the likely effects of different candidate treatments and select the one that is predicted to produce the best outcome. A digital twin that models the diet and metabolism of an individual would allow that person to understand how dietary and lifestyle choices can impact their health.  A digital twin that models the impact of disease on an individual's health would help them and their caregivers plan for future needs for disease management.

The application of digital twins to drug development is particularly important given the increasing time and costs of bringing new drugs to market. Drug development is a slow, resource-intensive process that places a substantial burden on patients in order to develop new therapies.  The planning and execution of clinical trials are a key component of this, and significant work needs to be done to make them faster and more beneficial for their participants.  Because clinical trials are well-controlled experiments in specific patient populations, they offer the opportunity to apply digital twins to provide concrete benefits for speed, statistical power, and decision making with well-controlled risks. 

Clinical trials produce broad assessments of the safety and efficacy of new treatments relative to placebo or standard of care. Drug safety and efficacy are inherently multidimensional concepts--i.e., drugs often affect complex systems in the body to alleviate multiple symptoms while potentially causing various side effects--which means that any computational models used to create digital twins of participants in drug trials need to be multivariate in order to produce forecasts for all outcomes of potential interest. In addition, the effect of a drug is inherently time-dependent, and participants in trials are typically assessed longitudinally across multiple time points. Taken together, these considerations imply that multivariate time-series models are needed to create digital twins of participants in drug trials; we call these models Digital Twin Generators (DTGs).

The availability of accurate DTGs would allow for the design of more efficient clinical studies while also unlocking new capabilities. For example, a participant's digital twin can be used to create something like an individual control group that addresses the question, ``What would likely happen to this participant if they were to receive a placebo or alternative treatment?''. With this knowledge, one could derive prognostic scores used to increase statistical power, estimate treatment effects down to the level of the individual participants in addition to population averages or subgroups, add simulated control groups to single-arm studies or during open-label extension periods, or even run simulated head-to-head comparisons of alternative treatments to assess comparative effectiveness.

Although the development of detailed mechanistic models is, at least in principle, one pathway to creating DTGs, the overwhelming complexity of the human body makes this a daunting task that is likely beyond the reach of current technologies and molecular-level datasets. An alternative approach is to apply Machine Learning (ML) techniques to train models as DTGs directly from phenotypic-level datasets. This requires the development of ML techniques that can learn to model stochastic, multivariate clinical time-series. 

In this work, we describe a novel generative neural network architecture that is capable of creating accurate DTGs in 13 different disease indications across 3 distinct therapeutic areas (neurodegeneration, immunology and inflammation, and general medicine) with no modifications other than hyperparameter tuning. This architecture combines a Neural Boltzmann Machine (NBM), an energy-based model with parameters set by neural networks, with deep neural networks designed to model multivariate time-series to create a highly flexible conditional generative model of clinical trajectories.  We train disease-specific DTGs based on this architecture using individual participant-level datasets aggregated from previously completed clinical trials, observational studies, and disease registries, assess their performance characteristics, and explore their interpretability through sensitivity analyses.

This report makes the following contributions:
\begin{itemize}
    \item We describe a novel generative neural network architecture for clinical trajectories based on an NBM that uses probabilistic modeling to address the intrinsic variability of clinical data.
    \item We discuss methodological innovations that allow these models to reliably train on heterogeneous clinical datasets with common issues like missing observations in order to produce well-calibrated probabilistic forecasts that capture complex temporal dependencies.
    \item We demonstrate that this architecture can produce accurate DTGs across many disease indications with no modifications other than hyperparameter tuning. 
    \item We report the generative performance of these models across a large and diverse clinical dataset covering 13 different disease indications, evaluating performance and model interpretability tools in each indication.
\end{itemize}

\begin{figure}
    \centering
    \includegraphics[width=0.85\linewidth]{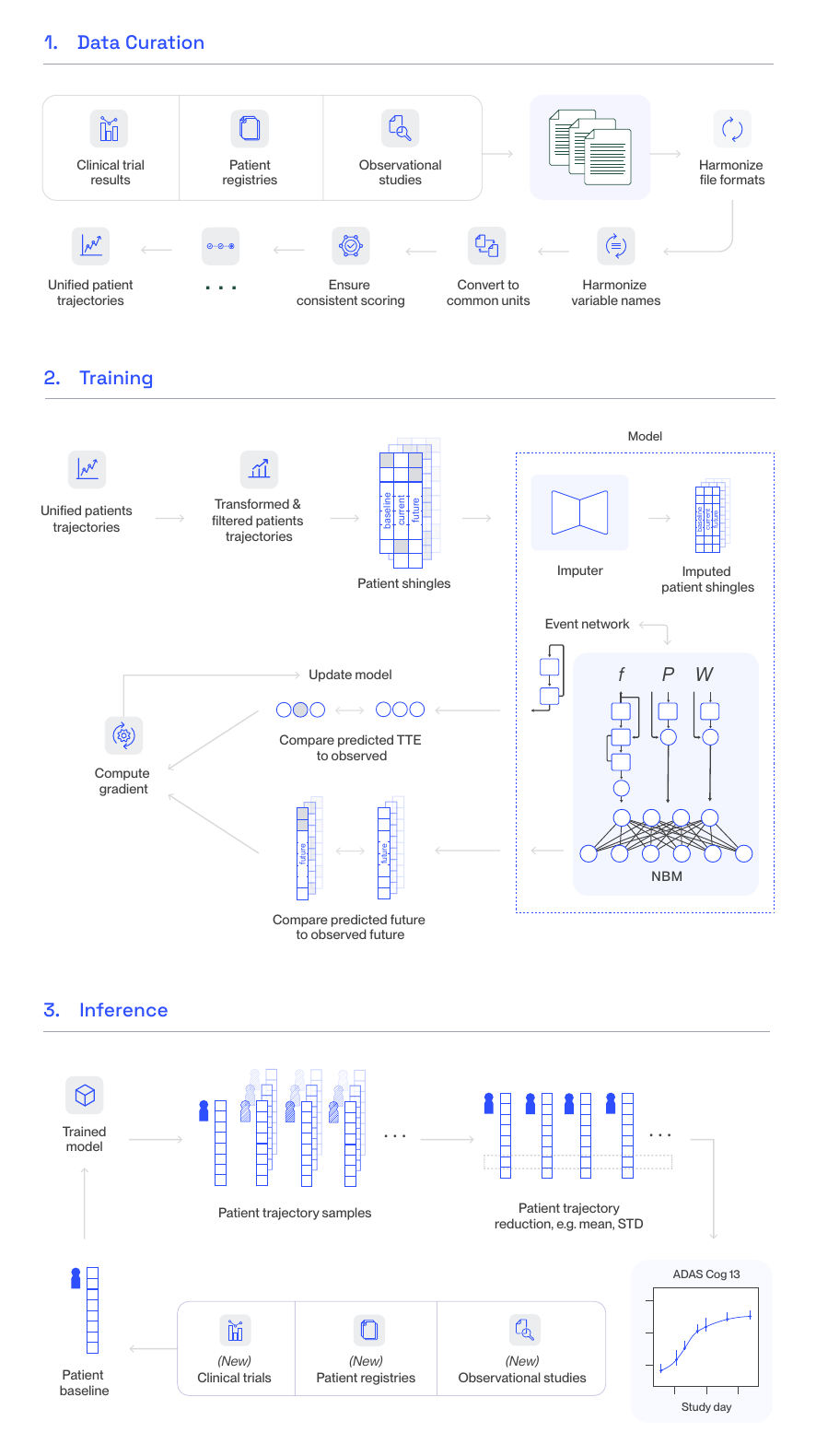}
    \caption{Overview of DTG creation and evaluation. There are three main phases: curation of data, training a DTG, and using the DTG to make digital twins for data outside the training set.}
    \label{fig:overview}
\end{figure}

An overview of the process used to develop and evaluate the DTGs is shown in \Fig{overview}.

In Section~\ref{sec:model} we present the NBM architecture that is used to produce a DTG in each disease indication. In Section~\ref{sec:applications} we discuss the datasets for each of the 13 disease indications, the hyper-parameters of the model and the process of selecting them and training the DTG. In Section~\ref{sec:validation} we show a broad set of metrics evaluating DTG performance.  In Section~\ref{sec:future} we discuss future directions for DTGs, and in Section~\ref{sec:conclusions} we conclude.

\section{Neural Boltzmann Machines for Generating Clinical Trajectories}
\label{sec:model}
Energy-based models provide a general framework for probabilistic modeling that can be viewed as an alternative to auto-regressive next token prediction, diffusion-based sample generation, or generative adversarial networks for efficiently learning high-dimensional joint distributions. General energy-based models can be highly expressive but are difficult to sample from, whereas specific architectures like Restricted Boltzmann Machines (RBMs) are relatively easy to sample from but have limited capacity. \emph{Neural
Boltzmann Machines} (NBMs)~\cite{llf2023} combine the two approaches to create an energy-based  model that is both highly expressive and relatively easy to sample from. 
An NBM is a neural network that models an observed state $\mathbf{y}$, and a latent state
$\mathbf{h}$ conditioned on some context $\mathbf{x}$ with a joint distribution
of the form
\begin{equation}
    p\left(\mathbf{y}, \mathbf{h} | \mathbf{x}\right) = 
    \mathcal{Z}^{-1}e^{-\mathcal{U}\left(\mathbf{y},\mathbf{h}|\mathbf{x}\right)},
\end{equation}
where $\mathcal{U}\left(\mathbf{y},\mathbf{h}|\mathbf{x}\right)$ is a joint energy function
parameterized by some neural network, which we will describe subsequently, 
and $\mathcal{Z}$ is an intractable normalization term taken as the integral (or sum) over
all possible $\mathbf{y}$ and $\mathbf{h}$.

By coupling the visible state to a latent state as in an RBM, an NBM provides a flexible and expressive model for the marginal distribution for the observed state $\mathbf{y}$ while enabling efficient block Gibbs sampling. We define the NBM's joint energy as
\begin{equation}
    \mathcal{U}\left(\mathbf{y}, \mathbf{h} | \mathbf{x} \right) 
    = 
    \frac{1}{2}\left[\mathbf{y}-\mathbf{f(\mathbf{x})}\right]^{T}{\rm P}(\mathbf{x})\left[\mathbf{y}-\mathbf{f(\mathbf{x})}\right]
    -
    \left[\mathbf{y}-\mathbf{f(\mathbf{x})}\right]^{T}{\rm W}(\mathbf{x})\mathbf{h}
    -\mathbf{b}^T(x) \mathbf{h},
\end{equation}
where the functions $\mathbf{f\left(\mathbf{x}\right)}$, $\mathbf{b\left(\mathbf{x}\right)}$, ${\rm P}(\mathbf{x})$,  ${\rm W}(\mathbf{x})$ are neural networks that return either a 
vector ($\mathbf{f}$ and $\mathbf{b}$), or a matrix (${\rm P}, {\rm W}$) output. In principle, 
the variables in the visible state may be either continuous or discrete, as in~\cite{llf2023}, but in this work we will focus on the case in which they are continuous. Likewise, we focus on the case in which the model has Ising latent variables, i.e. $h_i \in \{-1, +1\}$ and we set the bias on the hidden units to zero for simplicity (i.e., $\mathbf{b}(\mathbf{x}) = \mathbf{0})$. 
Parameterizing the energy with neural networks in this way is one of the 
principle innovations of the NBM, providing much more expressive power than typical RBMs that share this general form~\cite{llf2023}. 
Parameterizing distributional parameters with neural networks is a strategy employed 
in other types of models, notably in amortized variational inference~\cite{AVI}, of 
which the variational auto-encoder~\cite{VAE} is most famous.

By marginalizing the joint distribution over the latent state, we can derive the marginal probability distribution of the observed state conditioned on the context using
$p\left(\mathbf{y}|\mathbf{x}\right)=\sum_{H}p\left(\mathbf{y},\mathbf{h}|\mathbf{x}\right)$. 
In turn, this allows us to express a marginal energy function for the observed state, 
which we will need in order to train the model,
\begin{equation}
    \mathcal{U}\left(\mathbf{y}|\mathbf{x}\right)
    =
    \frac{1}{2}
    \left[\mathbf{y}-\mathbf{f(\mathbf{x})}\right]^{T}{\rm P}(\mathbf{x})\left[\mathbf{y}-\mathbf{f(\mathbf{x})}\right]
    -
    \mathbf{1}^{T} \left[ 
        \log\cosh\left( {\rm W}(\mathbf{x})\left(\mathbf{y}-\mathbf{f(\mathbf{x})}\right)\right)
    \right],
\end{equation}
where $\mathbf{1}$ is simply the all ones vector.
For a given architectural choice of $\mathbf{f}$, ${\rm P}$, and ${\rm W}$, 
training an NBM proceeds in a fashion very similar to a traditional
RBM, with the exception that gradients must be backpropagated through the constituent neural networks.  For a detailed description of how the NBM update gradients are calculated, we refer the
reader to our earlier work on the topic~\cite{llf2023}.

Up to now, time dependence of the state or context has been suppressed for clarity. 
To express the time-dependent observed state distribution required for longitudinally 
modeling clinical trajectories, we allow the 
neural networks 
$\mathbf{f}(\mathbf{x}, t)$, 
${\rm P}(\mathbf{x}, t)$, 
${\rm W}(\mathbf{x}, t)$ to depend on time $t$. 
Furthermore, we allow the state of the model at some future measurement time to be
conditioned on the state of the model at the current time, forming an auto-regressive link
between times in the joint distribution
\begin{equation}
     p(\mathbf{y}(t_{\text{future}}), \mathbf{h}(t_{\text{future}})\;\big|\; 
    \mathbf{x} = [\mathbf{y}(t_{\text{current}}), \mathbf{c}]).
\end{equation}
This construction allows us to begin with a measurement of a patient at an initial time point
$\mathbf{y}(0)$ (e.g., the initial visit in a clinical trial) and some time-independent context variables $\mathbf{c}$, then predict a future visit at some arbitrary time later. In principle, the conditioning set could include all previous time points if desired. During training, the model is fed a sequence of next-step ahead pairs, 
however, unlike discrete-time sequence modeling as in language models, 
the values of $t_{\text{current}}$ and $t_{\text{future}}$ may take any continuous positive value.
The need to model variable time steps arises from 
\emph{i}) the heterogeneous mixture of clinical trials and observational studies, which collected assessments at various intervals, we aggregated to create the training dataset, 
and \emph{ii}) because we desire to create digital twins that can forecast symptoms at \emph{any} time point in the future -- not just on a discrete, fixed grid.

\subsection{NBM Network Architecture for Disease Modeling}
\label{sec:sub-neural-arch}

An NBM is a generic, flexible, energy-based model that derives most of its behavior from its constituent neural networks. Therefore, creating a foundational architecture that can be used to train DTGs in a variety of indications comes down to making selections for the neural networks $\mathbf{f}$, ${\rm P}$, and ${\rm W}$. Good choices for these networks need to be expressive enough to accurately model multivariate clinical trajectories across
many diseases while also not being prone to overfitting to 
confounding factors when data availability is low. In addition, we use modular architectures for these constituent neural networks to aid in interpretability.  

The neural networks that parameterize the NBM energy at  $t_{\text{future}}$ serve different roles: although the mean of the observed distribution is not available in closed-form, it is heavily influenced by $f(\mathbf{x}, t_{\text{future}}, t_{\text{current}})$, the diagonal matrix $P(\mathbf{x}, t_{\text{future}}, t_{\text{current}})$ is similar to a precision matrix and heavily influences the variance of the observed distribution, and the matrix $W(\mathbf{x}, t_{\text{future}}, t_{\text{current}})$ projects the hidden state from a dimension M to the size of the observed state N and heavily influences the correlations between observed variables.

Due to the bipartite nature of the graph connecting the visible and hidden units, RBMs and NBMs can be sampled relatively efficiently by block Gibbs sampling. However, it's well known that the resulting Markov chains may mix slowly relative to the volume of phase space, so that it takes a long time to obtain a representative sample from the distribution defined by an RBM. Since NBMs are conditional generative models, it's not necessary to sample the same volume as an unconditional model, which makes representative sampling with a Markov chain easier. In addition, one can further speed up sampling based on the intuition that the mean (i.e., close to $f(\mathbf{x}, t_{\text{future}}, t_{\text{current}})$) is a high probability state and a good place to start the Markov chain \cite{llf2023}.

\begin{figure}
    \centering
    \includegraphics[width=0.9\linewidth]{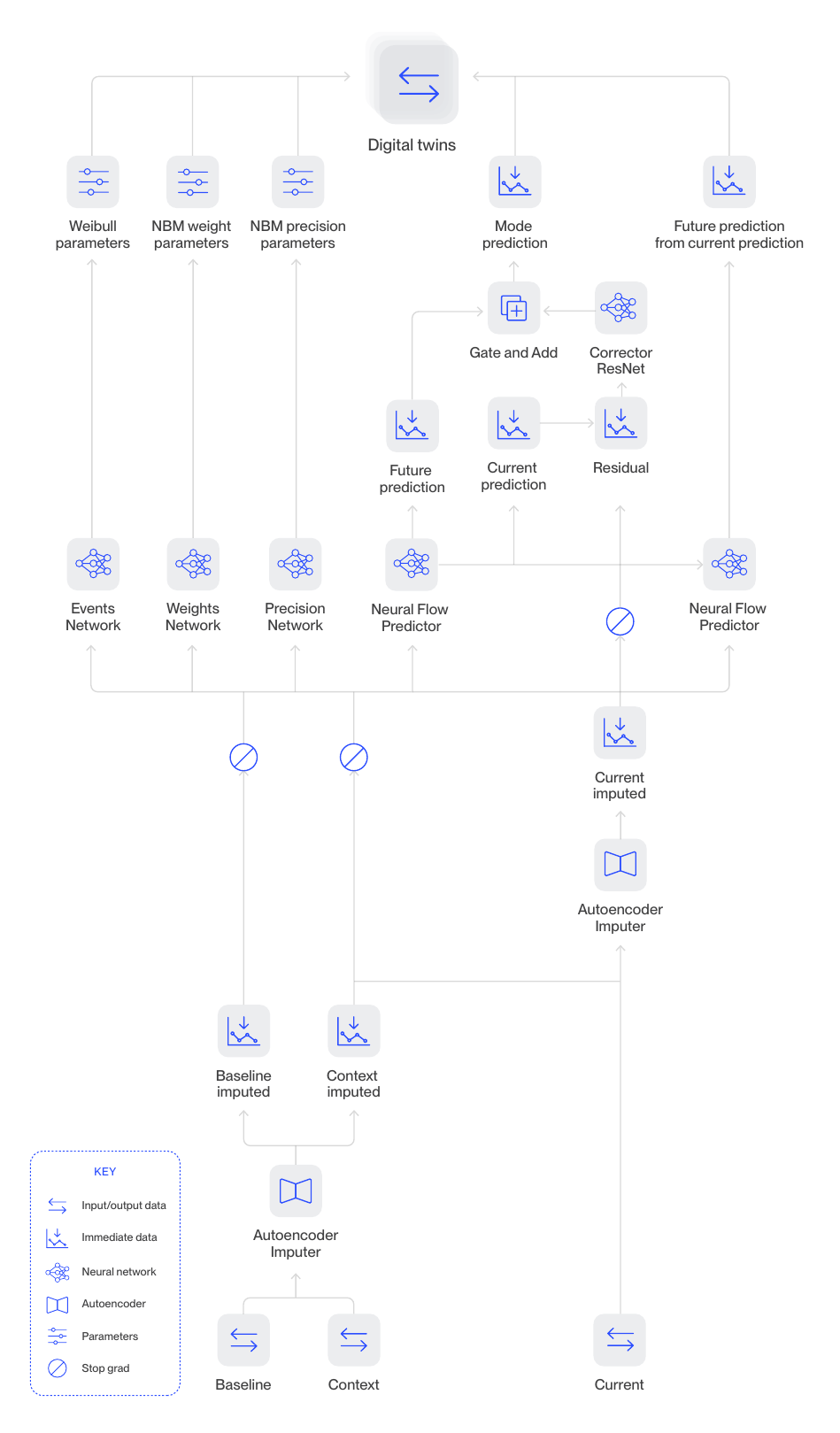}
    \caption{Overview of Neural Boltzmann Machine Architecture for Digital Twin Generators.  Individual networks, parameters, and outputs are described in the text in Section~\ref{sec:model}.}
    \label{fig:architecture}
\end{figure}

In \Fig{architecture}, we provide an overview of the architecture of the Neural Boltzmann Machine and its application to the Digital Twin Generators discussed in this section.

\paragraph{Imputation Network.} 
One challenge in training models on clinical data is that there are often many missing inputs. 
The neural networks in $\mathbf{f}$ , $P$, and $W$ are not themselves equipped to handle missing data in their inputs. To address this we use an auto-encoder network $\text{AEImputer}(\mathbf{x})$ to impute missing values into the data before they are passed to subsequent networks. The imputer first zero-fills the missing data before transforming it into an equal-shaped representation $\mathbf{y}$ that is compared to the non-missing values in a separate reconstruction loss.  The architecture uses a classic encoder-decoder design with encoder $e(\mathbf{x})$ and decoder architecture $d(\mathbf{x})$ taking the form:
\begin{align}
    l\left(\mathbf{x}\right) &= \phi \left(\text{Linear}\left(\frac{\text{LayerNorm}(\mathbf{x})}{\sqrt{\text{dim}(\mathbf{x})}}\right)\right) \\
    e(\mathbf{x}) &=  \left(l_{n} \circ \cdots \circ l_{1}\right)\left(\text{ZeroFill}(\mathbf{x})\right)\\
    d(\mathbf{x}) &= \left(\tilde{l}_{m} \circ \cdots \circ \tilde{l}_{1}\right)(\mathbf{x})\\
    \text{AEImputer}\left(\mathbf{x}=\left[\mathbf{y}, \mathbf{c}\right]\right) &= \text{where}\left(\mathbf{x} \text{ is present}, \mathbf{x}, d(e(\mathbf{x}))\right)
\end{align}
where $\phi(\cdot)$ is a non-linear activation function that we set to $\phi(\cdot) = \text{ArcSinh}(\cdot)$. Due to the conditional replacement at the end, the imputer only inserts values where there are none available in the data. Note that the imputer network does not have any dependence on time; it acts on each visit $\mathbf{y}$ separately. For the models we describe here, both the encoder and decoder employ two layers each. In addition, we detach the gradients from the imputed data before passing them to the rest of the network.

\paragraph{Mean Prediction Network.}
In many medical applications, it's common to need to predict many future observations based on a set of measurements taken at a single time point (along with other contextual information). For example, in a clinical trial, one is interested in forecasting the future evolution of a participant's clinical assessments from a set of baseline measurements taken during their first visit as part of the study. As a result, the problem does not look like a sequence-to-sequence prediction task but is instead a vector-to-sequence prediction task similar to a neural ordinary differential equation (Neural ODE). 

As a result, we specify $\mathbf{f}(\mathbf{x}, t_{\text{future}}, t_{\text{current}})$ to have a predictor-corrector design inspired by ODE solvers.
Predictions of the future time step are a combination of a prediction from
information available about a patient at an initial time point (e.g., trial randomization) and a prediction from the previous available time step. 
The prediction from the initial time point to a given time is handled by a 
prediction network, $g(\cdot)$ inspired by Neural Flows~\cite{bilovs2021neural},
which consists of a stack of residual layers,
\begin{align}
 g(\mathbf{y}(0), \mathbf{c}, t; \mathbf{s}) 
 &= \text{FlowBlock}\left[ 
        \mathbf{z}^{(L-1)}, \mathbf{c}, t, \mathbf{s}
     \right] \circ \cdots \circ \text{FlowBlock}\left[ 
        \mathbf{y}^{(0)}, \mathbf{c}, t, \mathbf{s}
     \right] \label{g-definition}
\end{align}
where $l$ is a layer index, $\mathbf{s}$ are a set of trained time-scale parameters
of the same dimensionality as $\mathbf{y}(0)$, each $\text{FlowBlock}$ outputs a tensor $\mathbf{z}^{l}$ of size $\mathbf{y}(0)$ and the $\text{FlowBlock}$ is 
defined as
\begin{align}
    \text{Norm}(\mathbf{x}) &= \frac{\text{LayerNorm}(\mathbf{x})}{\sqrt{\text{dim}(\mathbf{x})}} \\
\text{FlowBlock}\left[ \mathbf{x}, \mathbf{c}, t, \mathbf{s}\right] &=
    \theta \left(\mathbf{x} 
    + t \cdot \text{Linear}\left[ \text{Norm}\left(\left[ \mathbf{x}, \mathbf{c}\right]\right) \right]
    \right),
\end{align}
where $\theta(\cdot) = \mathbf{s} \odot \text{ArcSinh}\left(\mathbf{s}^{-1} \odot \cdot \right)$ is a non-linearity with trainable parameters $\mathbf{s}$, and $N$ is number of elements in $\mathbf{y}(0)$. Note that, unlike the original Neural Flows implementation, we do not enforce the invertibility of these flow block networks. 
In our experiments, we find $L = 3$ to be a good selection for the depth of the 
prediction network.
The prediction network is used to produce predictions at both $t_{\text{future}}$
and $t_{\text{current}}$, however, its predictions for these
two time points are made independently -- they both are generated conditioned
entirely on initial conditions $\mathbf{y}(0)$ and time-independent context 
$\mathbf{c}$. 

In order to introduce auto-correlation through time and permit sensible autoregressive 
sampling conditioned on previous predictions, 
the residual between the output of the prediction network at $t_{\text{current}}$
and the currently available observed state at $t_{\text{current}}$ is passed through
the \emph{corrector} network.
The corrector network may be a deep multilayer perceptron, and its output is combined with 
the output of the prediction network at $t_{\text{future}}$
to form the final output for $f$,
\begin{align}
    q(\cdot) &= \phi\left(\text{Linear}\left[\cdot\right]\right) \\
    f(\mathbf{x}, t_{\text{future}}, t_{\text{current}}) &= g(\mathbf{x}, t_{\text{future}}) + e^{-\boldsymbol{\lambda} \left(t_{\text{future}}-t_{\text{current}}\right)} \left(q_{N} \circ \cdots \circ q_{1}\right)\left( g(\mathbf{x}, t_{\text{current}}) - \mathbf{y}(t_{\text{current}}) \right)
\end{align}
In the experiments we present, we employ a single layer for $q(\cdot)$ with an identity activation function $\phi$. Although the predictor-corrector architecture may seem mysterious at first glance, the intuition is relatively easy to grasp. We use a neural network to make a prediction for the state of the system at time $t_{\text{current}}$ and measure the residual error. Next, we use the neural network to make a prediction for the state of the system at a future time $t_{\text{future}}$. If the difference between the two times is short, then we expect the residual errors for the predictions at the two times to be similar. On the other hand, if the difference between the two times is long, then knowing the residual error at the first time doesn't help us predict what it will be at the next time. 

As an additional regularization, a secondary prediction at $t_{\text{future}}$ 
is made, but using $t_{\text{current}}$ as an initial condition instead of 
$\mathbf{y}(0)$,
\begin{align}
    f^{*}(\mathbf{x}, t_{\text{future}}, t_{\text{current}}) 
    &= 
    g([g(\mathbf{x}, t_{\text{current}}), \mathbf{c}], t_{\text{future}}) \label{composed-g}.
\end{align}
We aim to have the difference between $f$ and $f^{*}$ minimized, 
akin to the requirement for an autonomous ODE. 
The difference between these two predictions is passed to a consistency 
loss discussed in the next section. A similar strategy is used in the original formulation of the Neural Flows and in Consistency Models~\cite{ConsistencyModel}.

\paragraph{Weight Prediction Network.}
At the heart of $W(\mathbf{x}, t_{\text{future}}, t_{\text{current}})$ is a matrix-valued function $w(\mathbf{x})$, meaning it outputs a matrix of size $N \times M$, in order to produce a linear projection of the latent state $\mathbf{h}$ to the same size as the observed state $\mathbf{y}$. Time dependence of this function is obtained via multiplication by a time-dependent gate, which is a common theme among all the component functions here:
\begin{align}
m(\mathbf{x}) &= \phi\left(\text{Linear}\left[\frac{\text{LayerNorm}(\mathbf{x})}{\sqrt{\text{dim}\left(\mathbf{x}\right)}}\right]\right) \label{eq:simple-layer} \\
w(\mathbf{x}) &= \frac{1}{\sqrt{\text{output dim}}}\left(m_{N}\circ \cdots \circ m_{1}\right)(\mathbf{x}) \\
    {\rm W}(\mathbf{x}, t_{\text{future}}, t_{\text{current}}) &= e^{-\boldsymbol{\lambda}\left(t_{\text{future}}-t_{\text{current}}\right)}w(\mathbf{x})
\end{align}
Where in our experiments presented here the number of layers in $w(\mathbf{x})$ was 1 and $\phi(\mathbf{x})$ was the identity function.

\paragraph{Precision Prediction Network.} 
The precision function $P(\mathbf{x}, t_{\text{next}}, t_{\text{current}})$ is used to scale the observed state within the NBM's energy function. In this work, it is comprised of a network $p(\mathbf{x}, t_{\text{next}})$ and a gate function:
\begin{align}
p(\mathbf{x}) &= \left(m_{N} \circ \cdots \circ m_{1}\right)(\mathbf{x}) \\
    {\rm P}(\mathbf{x}, t_{\text{next}}, t_{\text{current}}) &= \exp \left[\boldsymbol{\beta} - \log \left(1 + (1-e^{-\left|\boldsymbol{\lambda}\right|\left(t_{\text{next}}-t_{\text{current}}\right)}) e^{p(\mathbf{x}, t_{\text{next}})}\right)\right]
\end{align}
Where bold Greek symbols represent learnable tensors of appropriate size, $m(\mathbf{x})$ is defined in equation  \ref{eq:simple-layer}, and in experiments presented here the number of layers was 1 with the activation set to the identity function. This particular choice for the scaling of the precision matrix with the time difference can be derived by approximating the residuals with an Ornstein-Uhlenbeck process, but we leave this as an exercise for the adventurous reader. 

\subsection{Time-to-Event Modeling}
\label{sec:sub-tte}

So far, we've considered the task of forecasting the time evolution of some clinical variables that we may imagine evolving according to some unknown stochastic differential equation; that is, they are continuously evolving in time. However, there are many cases in a medical context for which we need to forecast the time when a particular discrete event will take place. Although one could, at least in principle, forecast such an event as the first passage time of a continuously evolving variable, this approach doesn't work well in practice, and it's easier to incorporate dedicated networks to predict specific events. 

As a result, our experiments use a bolt-on approach that connects a Time-to-Event (TTE) component model to the imputer module of the overall model, but that is disconnected from the longitudinal predictions. Using this approach, we provide a distribution for the time it takes to see a particular event, conditioned on information available at the initial time point. The distribution we use is a univariate Weibull distribution. Weibull distributions are commonly employed to model time-to-event or time-to-failure analyses (known as accelerated failure time models) \cite{liu2023using}. In particular, we model the log of the time-to-event, which transforms the Weibull to a Gumbel distribution whose two scalar parameters are provided by a scalar-valued neural network $a(\mathbf{x})$ and learnable scalar tensor $\sigma$:
\begin{align}
    r(\mathbf{x}) &= \mathbf{x} + \phi\left(\text{Linear}\left[\frac{\text{LayerNorm}(\mathbf{x})}{\sqrt{\text{dim}(\mathbf{x})}}\right]\right) \label{tte-distribution}\\
    a(\mathbf{x}) &= \text{Linear}\left[\frac{\text{LayerNorm}\left(r_{N} \circ \cdots \circ r_{1}\right)(\mathbf{x})}{\sqrt{\text{dim}(\mathbf{x})}}\right]\\ 
    \log\left(T\right) &= a(\mathbf{x}) + \sigma \times \varepsilon \\ 
    \varepsilon &\sim \text{Gumbel}\left[0, 1\right]
\end{align}
where $\phi(\cdot)$ is a nonlinearity that we set to $\text{ArcSinh}(\cdot)$ as in previous networks, $a(\mathbf{x})$ is comprised of a stack of residual blocks $r(\mathbf{x})$ followed by a LayerNorm and Linear layer to collapse the intermediate (shaped like the input  $\mathbf{x}=\left[\mathbf{y}\left(0\right), \mathbf{c}\right]$) to a scalar.  The residual block is comprised of LayerNorm, scaling layer, and Linear layer followed by an ArcSinh activation with a skip connection from end to end. In experiments presented here, the number of residual layers was 1. In principle, one could add additional residual blocks to this network if required.

\subsection{Why So Many LayerNorms?}

We use a relatively unusual pattern in which the inputs to all of these neural networks are wrapped inside a function,
\begin{equation}
\text{Norm}(\mathbf{x}) = \frac{\text{LayerNorm}(\mathbf{x})}{\sqrt{\text{dim}(\mathbf{x})}} 
\end{equation}
Applying this normalization to the inputs of a linear layer is a simple way to approximately control the variance of the outputs of that layer, which will be determined by the $L_2$ norm of the layer's weights. Since we are using $L_2$ weight decay to regularize the parameters in the networks, applying this normalization helps to put all of the network's parameters on the same scale, making it easier to tune the weight decay.

\subsection{Training Losses and Regularizations}

Energy-based models like NBMs are typically trained via approximate maximum likelihood. In the case of RBMs and NBMs, this uses an algorithm known as Contrastive Divergence. In our case, however, we use a more complex loss function that also includes terms to train the autoencoder for imputation and the time-to-event network, as well as to regularize training overall. All of the network parameters are trained simultaneously by differentiating through a loss function that is a weighted sum of the components outlined below. The weights of these components are hyper-parameters of the training procedure.

\paragraph{Reconstruction Loss.} This loss trains the
AEImputer to accurately impute missing data points 
by minimizing the mean squared error between the original data.
Notably, the AEImputer is trained to impute \emph{both} 
missing contextual information as well as longitudinal data. 
It is also time-invariant and is trained on available 
patient visits within the training batch.
Additionally, every visit sample is assigned a weight, $w_i$, to make 
the \emph{per-patient} loss contribution uniform.
Assuming a zero-filled sample, $\mathbf{x}$, the imputer 
loss is defined as
%
\begin{align}
L_{\text{imputer}} &=
    \frac{1}{VN} \sum_{i=1}^V
    \sum_{j=1}^{N}
    w_i \cdot m_{i,j} \cdot \left| 
        \text{AEImptuer}(\mathbf{x}_i)_j - \mathbf{x}_{i,j}
    \right|^2
\end{align}
where $V$ is the number of visit samples used to update the AEImputer,
and $m_{i,j}$ is a masking coefficient with value $0$ if variable $j$ was missing from
$\mathbf{x}_{i}$ and $1$ otherwise. 
Lastly, in order to prevent overfitting NBM longitudinal predictions 
to potentially confounding patterns of missingness from multi-source 
training data, $L_{\text{imputer}}$ is optimized independently
via stop-gradient -- in this way, the AEImputer is \emph{co-trained} with the 
rest of the NBM, rather than having gradients flow fully end-to-end.

\paragraph{Contrastive Divergence.} As in an RBM, the main contribution to the
training of an NBM is an approximate gradient of the negative log-likelihood computed via
contrastive divergence. 
It includes gradients from both the actual data
points (positive phase) and the model-generated data samples (negative phase).
The gradients can be computed by differentiating the following loss function,
\begin{align}
    L_{\text{RBM}} &= \mathbb{E}_{\text{batch}}\left[U(\mathbf{y}|\mathbf{x})\right] - \mathbb{E}_{\text{batch}}\left[\mathbb{E}_{\mathbf{y}|\mathbf{x}}\left[U(\mathbf{y}|\mathbf{x})\right]\right].
\end{align}

\paragraph{Feature-wise MSE Loss.}

This loss assesses the fidelity of the prediction network up to the 
predicted precision of each longitudinal variable by comparing the 
output of $\mathbf{g}(\cdot)$ at a time-point $t_{\text{next}}$ to its ground-truth
observation. After first performing a zero-filling to account for missing entries in
$\mathbf{y}(t_{\text{next}})$, the loss is calculated as

\begin{align}
L_{\text{featurewise\_mse}} &= 
    \mathbb{E}_{\text{batch}}\left[ 
    \sum_{j = 1}^{N}  m_j \cdot  P_j \cdot 
    \left| g(\mathbf{y}(0), \mathbf{c}, t_{\text{next}})_{j}  - y_{j}(t_{\text{next}})\right|^2 
    \right],
\end{align}
where $m_j$ is a masking coefficient with value $0$ if variable $j$ was missing from
$\mathbf{y}(t_{\text{next}})$ and $1$ otherwise.

\paragraph{Consistency Loss.} Ensures that
the model’s predictions are consistent over different forecasting intervals,
enhancing the temporal coherence of the model's outputs. This term is
computed as the expectation of the square difference between 
Eq.~\ref{composed-g} and Eq.~\ref{g-definition},
\begin{align}
    L_{\text{consistency}} 
    &= \mathbb{E}_{\text{batch}}\left[
        \left|
            g(\mathbf{y}(0), \mathbf{c}, t_{\text{next}})) 
            - g(g(\mathbf{y}(0), \mathbf{c}, t_{\text{current}}), \mathbf{c}, t_{\text{next}})
        \right|^{2}
    \right].
\end{align}

\paragraph{Event Loss.} Quantifies the accuracy
of event time predictions, relevant in models that feature a time-to-event
prediction. The loss is the log likelihood of the observed event time under
a Weibull distribution parameterized as shown in 
Eq.~\ref{tte-distribution}, which is inspired by \cite{liu2023using}.

\label{sec:training}

In addition to the above terms, we apply $L_2$ weight decay to the model parameters. We now turn from a general discussion of the architecture to details about how the model is trained and applied to individual disease indications.

\subsection{Training Strategy}
\label{sec:training-overview}

We use $k$-step Contrastive Divergence (CD) \cite{hinton2002training,
tramel2018deterministic} with additional penalties to train NBM models. This
objective permits us to use standard backpropagation to compute gradients for
the parameters of the neural networks involved in $f(\mathbf{x})$,
$P(\mathbf{x})$, and $W(\mathbf{x})$. Parameter optimization is done with
AdamW~\cite{adamWpaper} acting on minibatches of participant trajectories.
Samples of the observed state $\mathbf{y}$ for each observation in the batch are
computed using $k$ MCMC steps (the last sample is taken) without tracking the
gradient. These samples for $\mathbf{y}$ are then used to compute an empirical
expectation over the differentiable marginal energy $U(\mathbf{y}|\mathbf{x})$
in the negative phase of the contrastive divergence loss. The positive phase of
the contrastive divergence, which samples $\mathbf{y}$ from the data,  is
likewise differentiable. 

For each minibatch of patient trajectories, we convert the trajectory of patient
observations into a sequence of triplets comprised of the information at
the initial time point, the current visit, and a future visit. We compute all possible
causal triplets. This design enables the model to learn the progression of the
patient's state over different time intervals.

\section{Applying Neural Boltzmann Machines to Clinical Data}
\label{sec:applications}

\subsection{Clinical Trial Dataset Curation and Preprocessing}
\label{sec:data}

\begin{table}[H]
    \centering
    \begin{tabular}{llrr}
        \toprule
        \multirow{1}{*}{\textbf{Therapeutic Area}}  & \textbf{Disease Indication} & \textbf{Patients} & \textbf{Patient-Visits} \\
        \midrule
        \multirow{5}{*}{Neurodegeneration}          & Alzheimer's Disease (AD)            & 25,212  & 83,450  \\ 
                                                    & Huntington's Disease (HD)           & 13,286  & 33,323  \\
                                                    & Amyotrophic Lateral Sclerosis (ALS) & 10,887  & 83,581  \\ 
                                                    & Parkinson's Disease (PD)            & 2,199   & 9,401  \\ 
                                                    & Frontotemproal Dementia (FTLD)      & 1,412   & 4,182  \\
        \midrule
        \multirow{4}{*}{Immunology \& Inflammation} & Rheumatoid Arthritis (RA)           & 70,873  & 150,478  \\
                                                    & Psoriasis (PSO)                     & 6,202   & 15,215  \\
                                                    & Crohn's Disease (CD)                & 2,944   & 15,368  \\
                                                    & Atopic Dermatitis (ATD)             & 1,925   & 5,614  \\
        \midrule
        \multirow{4}{*}{General Medicine}           & Hypertension (HTN)                  & 8,955   & 32,349  \\
                                                    & Dyslipidemia (DLD)                  & 4,030   & 21,152  \\
                                                    & Type 2 Diabetes (TTD)               & 3,099   & 21,105  \\
                                                    & Acute Ischemic Stroke (STR)         & 2,613   & 10,087  \\
        \midrule 
        \multicolumn{2}{l}{\emph{Total}}                                                  & 153,637 & 485,305\\
        \bottomrule
    \end{tabular}
    \caption{
        The datasets used for training and evaluation of DTGs in each indication, in terms of the number of patients and patient-visits.
    }
    \label{tab:data-train}
\end{table}

We trained and evaluated DTGs in 13 different indications using datasets aggregated from a variety of sources, including previously completed clinical trials, observational studies, and registries. Across indications, data came from 156 clinical trials and 33 observational studies and registries, with study data totaling more than 400,000 patients with more than 3.8 million visits.  This data was was processed and harmonized into training datasets for each indication, with data filtered based on sample quality and relevance for common clinical trial outcomes.  The resulting datasets used for training and evaluation of the DTGs are described in \Tab{data-train}.

Each of these datasets contains multivariate clinical trajectories taken from individual study participants, where each participant was assessed in one or more follow-up visits at varying intervals. The sample sizes, measured in terms of the number of patients or the total number of patient visits (which is more appropriate for time series models), vary by roughly one order of magnitude across these indications. These datasets generally contain common clinical outcome assessments used in clinical studies for the given indications in addition to lab tests, data on biomarkers, and medical history. Of course, the particular variables are different for each indication. 

Clinical datasets are often heterogeneous, multimodal, and have varying data
quality standards. These problems are exacerbated in this case because these datasets do not come from individual studies, but rather were constructed by aggregating data from multiple studies in each indication. Thus, heterogeneity of datasets within a given indication may mean
that studies measure different sets of outcomes assessments or administer custom versions of assessments; studies may also have different durations of data collection, with varying visit frequencies, and may have enrolled different populations. Multimodal clinical datasets may include variables with different data types (i.e., continuous, binary, categorical, and survival). Assessments and variables may be recorded in the raw data with different, nonstandard conventions. Lastly, data quality standards may vary significantly across clinical datasets. Missing data, data entry errors, discrepancies between the raw data and data dictionaries, and any other inconsistencies need to be addressed during data harmonization
efforts.  

These characteristics of clinical data pose a significant challenge when trying
to harmonize them into single, integrated datasets for each indication. To accomplish this, we use configurable and standardized data processing pipelines to leverage software tools and domain knowledge to build high-quality datasets for training and validation. 
There are key steps to ensure good performance of the DTG when preparing data from any single data source: standardization of data with dimensions, standardized encoding  
of categorical or ordinal data, outlier detection for data with incorrect encoding, uninterpretable data,  
or unphysical measurements, and entity resolution for data such as medical histories or medication  
information.
These actions ensure data from each source is high quality while also remaining faithful to the original dataset.  There are also key steps performed as a result of harmonizing data across multiple studies, which are important to produce larger datasets that are consistent across studies and meet the need for a single, consistent data model for machine learning applications.  These steps primarily focus on transformations to data that remove inconsistencies in data-taking processes, for example, correcting supine vital sign measurements into the more common sitting version, or mapping data between versions of an assessment to ensure it can be consistently interpreted across studies.  In cases such as these, harmonization requires making choices that will make the data less specific and less faithful to individual studies in order to create a consistent, multi-study dataset.

\subsection{Hyper-parameters}
\label{sec:disease-specific}

We used the same NBM architecture to train DTGs in each of the 13 indications, but we did optimize model and training hyper-parameters for each indication. We found that the following hyper-parameters were most significant when attempting to tune a DTG to best fit the data in a given indication. 

\paragraph{AEImputer Embedding Dimension.} Each clinical dataset
has a different number of observations at the initial visit, both for the time-dependent variables as well as time-invariant variables that make up the context, such as labs, vitals, or medical history and genetic markers. Furthermore, each clinical dataset consists of a heterogeneous mix of data sources, each of which records different combinations of these variables at frequencies. Therefore, the longitudinal pattern of missingness is highly disease-dependent, making the tuning of the autoencoder-imputer network an important aspect of model performance. As a result, tuning the embedding dimension of the AEImputer is a necessary step to ensure a proper balance between predictive performance and generalization.

\paragraph{Weight Decay.} We apply $L_2$ weight decay to all of the parameters in the NBM to regularize the network and prevent overfitting. As mentioned previously, we applied a particular form of LayerNorm throughout the model in an effort to make all of the model parameters the same order of magnitude so that tuning the weight-decay penalty is easier. As a result, even though we separately tuned the weight-decay penalties for each sub-network, we found that the tuned penalties were all within the range from $[0.1, 0.5]$.

\paragraph{Block Gibbs Sampling Steps.} Samples are generated from an NBM by block Gibbs sampling. That is, one iterates between sampling from $\mathbf{y} \sim p(\mathbf{y} | \mathbf{h}, \mathbf{x})$ and $\mathbf{h} \sim p(\mathbf{h} | \mathbf{y}, \mathbf{x})$. We found that 16 or fewer iterations was generally appropriate for our disease-specific DTGs. Although, one can generally improve the performance by adding more iterations at the expense of increased computational cost. 

\paragraph{NBM Hidden Dimension Size.} The number of latent variables in the NBM (i.e., the dimension of $\mathbf{h}$) sets the capacity for the model to capture correlations between observed variables. Choosing too large of a hidden dimension balloons the size of the weights network, which increases the computational costs and may increase the likelihood for the network to overfit to the training set.  

\paragraph{Event Loss Weighting.} Only some of the indications we considered in our experiments included specific time-to-event variables. However, it was necessary to tune the weight applied to the time-to-event loss when such variables were present. Technically, the two networks are decoupled, but the relative weights of the loss functions could require tuning due to interactions with the learning rate of weight decay parameters. 

\vspace{5mm}

The model and training hyper-parameters were tuned to optimize the DTG's performance in each indication using 5-fold cross-validation. However, this tuning was largely performed by human evaluation because it's difficult to derive a single metric that captures the performance of a multivariate DTG. We considered various metrics such as mean-squared prediction errors for each variable at multiple time intervals, calibration of predicted quantiles at various time intervals, the ability to accurately model correlations and autocorrelations between variables, and others. Therefore, selection of ``the best'' hyperparameters for a given indication was admittedly subjective. 

In addition, each indication had a different number of time-dependent and time-independent context variables, which affects the number of parameters in the resulting model. The input and output dimensions as well as the number of model parameters for each indication are shown in Tab.~\ref{tab:model_params}. 

\begin{table}[ht]
\centering
\begin{tabular}{lrrr}
\toprule
Indication & Inputs &  Outputs & Parameters\\
\midrule
PD   & 112 & 91 & 221K \\
STR  & 110 & 58 & 204K \\
HD   & 80 & 57 & 144K \\
AD   & 76 & 54 & 130K \\
ALS  & 71 & 50 & 115K \\
RA   & 69 & 37 &  87K \\
CD   & 66 & 22 &  42K \\
TTD  & 42 & 14 &  21K \\
DLD  & 45 & 11 &  18K \\
FTLD & 34 & 20 &  11K \\
HTN  & 38 & 9 &   7K \\
PSO  & 42 & 4 &   7K \\
ATD  & 24 & 6 &   6K \\
\bottomrule
\end{tabular}
\caption{%
    Per-disease breakdown of the number of input baseline variables (\emph{Inputs}) 
    used to condition the generation of the disease-specific NBM, 
    the number of time-varying longitudinal variables predicted
    by the NBM (\emph{Outputs}),
    and the final trainable parameter count of the resulting NBM model,
    including all sub-networks of the NBM (\emph{Parameters}).
    \label{tab:model_params}
}
\end{table}

\section{Evaluation and Validation}
\label{sec:validation}

Good digital twins should be statistically indistinguishable from real observations of patients. Evaluating whether or not our digital twin generators are accurately modeling a given population of patients is a complex problem, because there are many ways to evaluate the goodness-of-fit for a generative model. For example, for a population, one can compare the moments of the data distribution to the distribution that the DTG returns (given some per-patient context information for the population). Of course, we are interested in modeling many potential populations, not just one in particular. We want the DTG to generalize and accurately model the distribution of any realistic population that we supply to it. 

We will employ a couple of tools to estimate generalization of the DTG. First, to prevent the model from overfitting on and potentially memorizing its training data, we use five-fold cross-validation to evaluate the models. This means that for each indication, the training dataset is split into five folds of approximately equal size. For each fold, predictions are generated from a model trained on the remaining four folds. Predictions from the five separate models evaluated on their held-out folds are merged and used to compute evaluation metrics reported in this section. Based on this validation method, we find that for all indications the same model architecture with minimal hyper-parameter tuning can learn to make accurate predictions across indications.

Another way to probe generalization is to examine the performance of the model for different, clinically relevant sub-populations or "cohorts". These cohorts are defined with some inclusion or exclusion criteria on the baseline features of the patients and so we expect the cohorts to have different clinical behavior over time. Likewise, one can also stratify a population on a single variable, which also creates cohorts of a different kind. Ideally, we will see that for each sub-population or patient stratification, the distribution over that sub-population matches the distribution that the DTG returns given that sub-population's baseline information.

\begin{figure}
    \centering
    \includegraphics[width=12cm, trim = 2cm 4cm 2cm 2cm, clip]{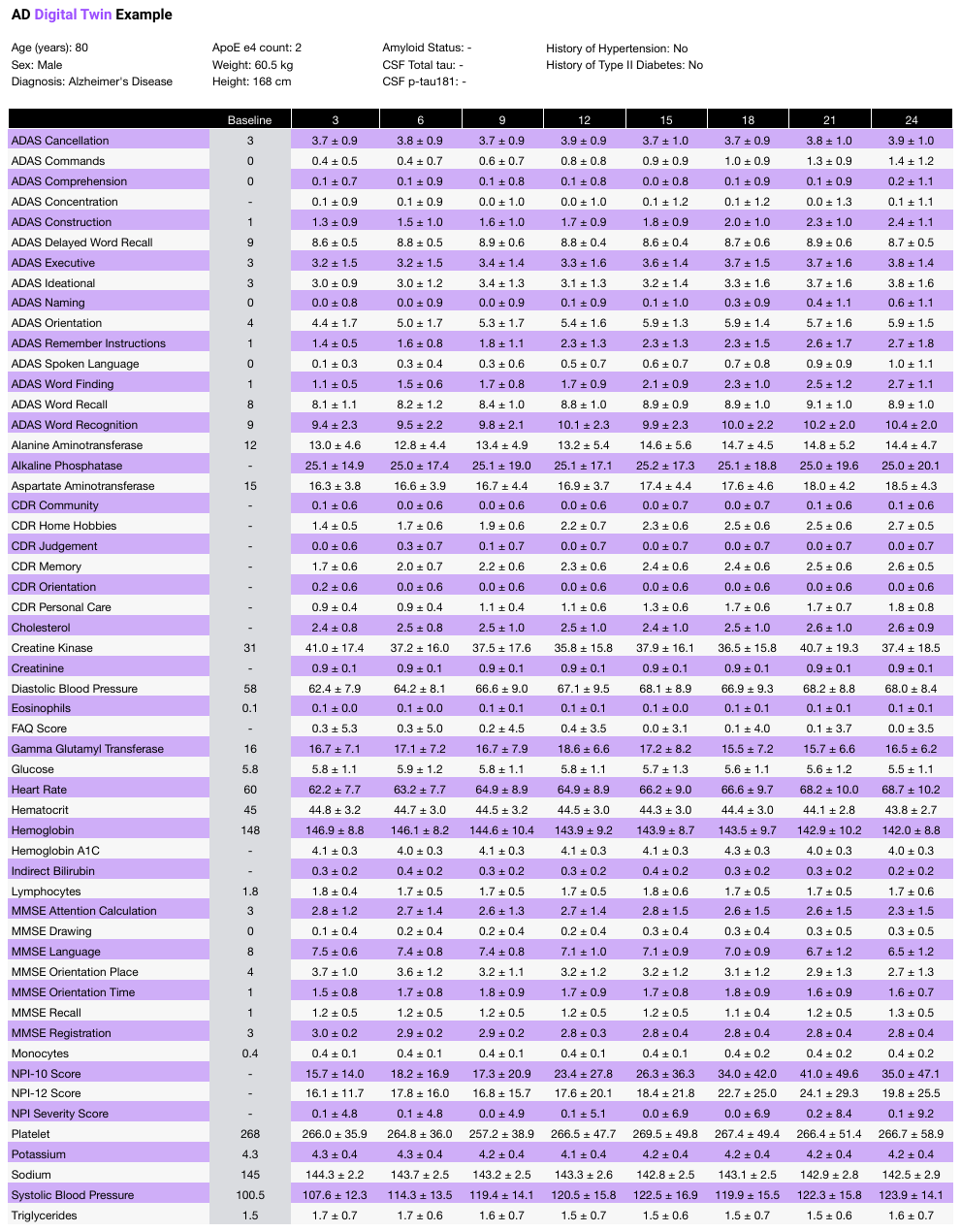}
    \caption{An example twin record for a single AD patient is shown, which reports the mean and standard deviation of the twins for that patient. A sample of static baseline information is given at the top of the record. A sample of longitudinal outcomes predicted by the model are supplied as rows, with columns indicating the time in months since baseline visit.}
    \label{fig:ad-example-twin}
\end{figure}

\begin{figure}
    \centering
    \includegraphics[width=12cm, trim = 2cm 8cm 2cm 2cm, clip]{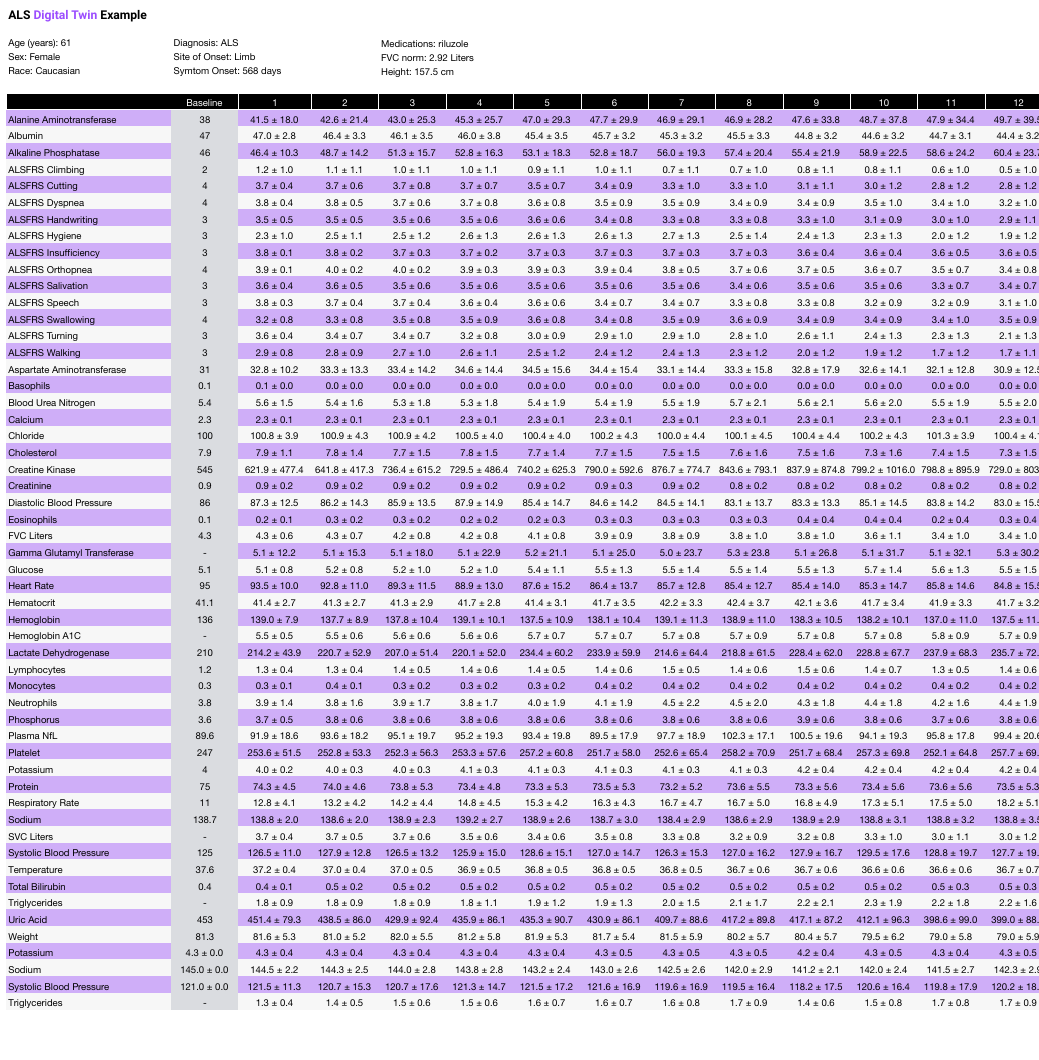}
    \caption{An example twin record for a single ALS patient is shown, which reports the mean and standard deviation of the twins for that patient. A sample of static baseline information is given at the top of the record. A sample of longitudinal outcomes predicted by the model are supplied as rows, with columns indicating the time in months since baseline visit.}
    \label{fig:als-example-twin}
\end{figure}

\begin{figure}
    \centering
    \includegraphics[width=12cm, trim = 2cm 19cm 2cm 2cm, clip]{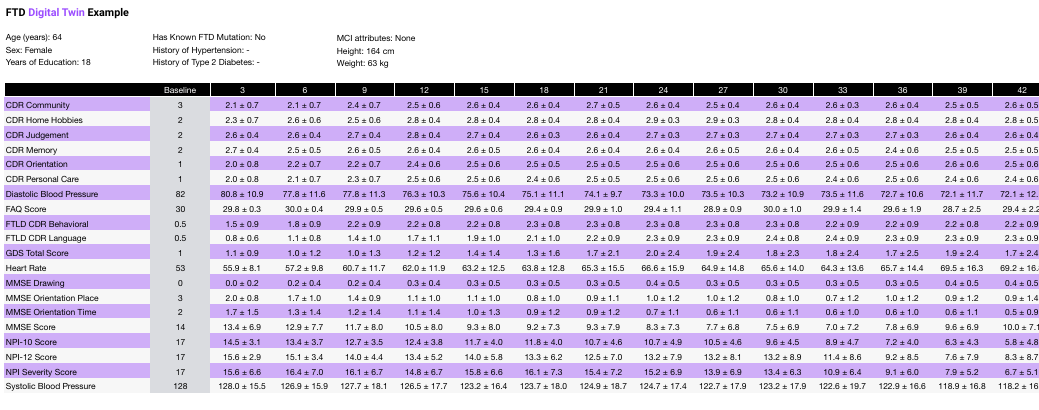}
    \caption{An example twin record for a single FTD patient is shown, which reports the mean and standard deviation of the twins for that patient. A sample of static baseline information is given at the top of the record. A sample of longitudinal outcomes predicted by the model are supplied as rows, with columns indicating the time in months since baseline visit.}
    \label{fig:ftd-example-twin}
\end{figure}

\begin{figure}
    \centering
    \includegraphics[width=12cm, trim = 2cm 2cm 2cm 2cm, clip]{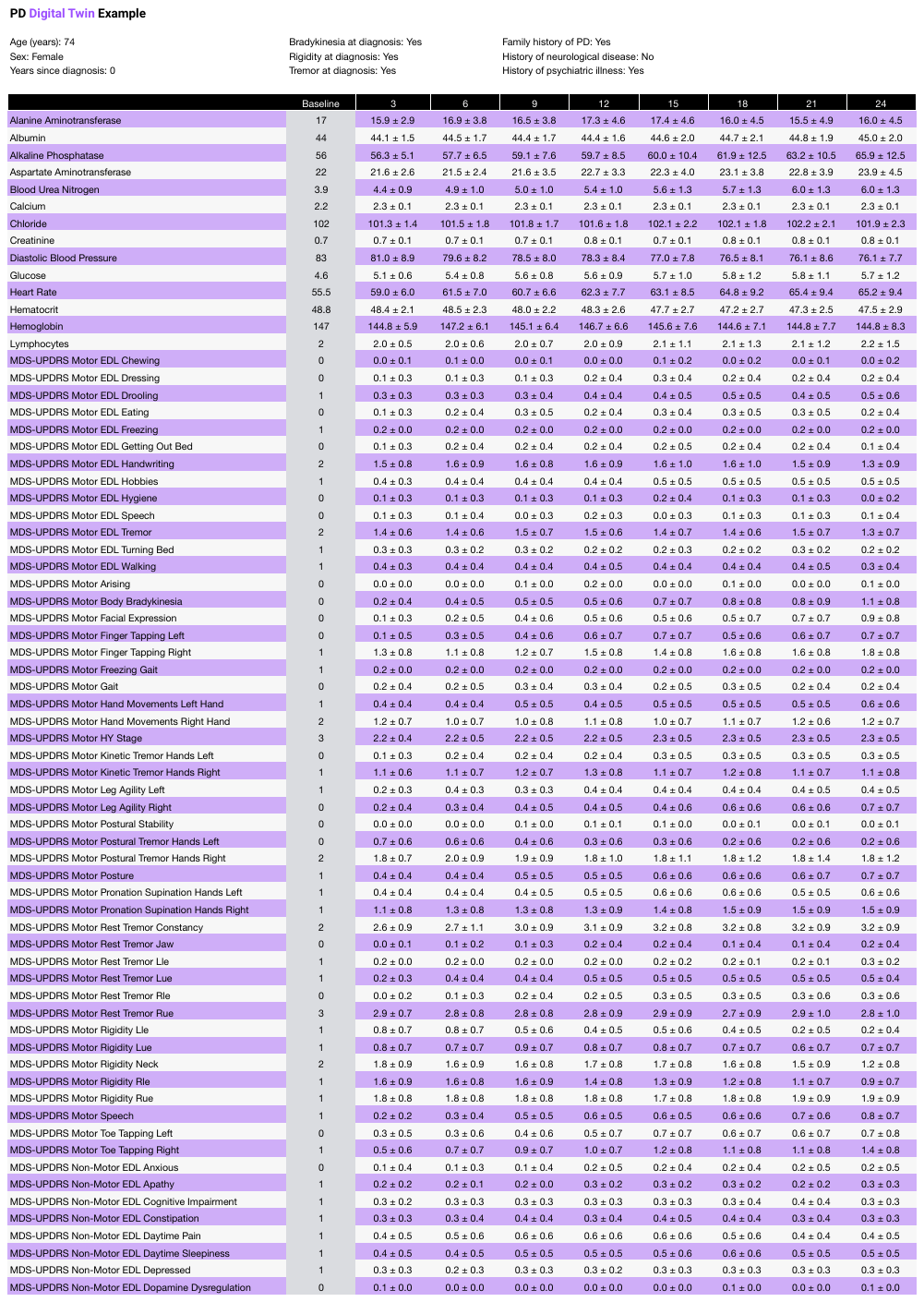}
    \caption{An example twin record for a single PD patient is shown, which reports the mean and standard deviation of the twins for that patient. A sample of static baseline information is given at the top of the record. A sample of longitudinal outcomes predicted by the model are supplied as rows, with columns indicating the time in months since baseline visit. The number of longitudinal records have been truncated.}
    \label{fig:pd-example-twin}
\end{figure}

\begin{figure}
    \centering
    \includegraphics[width=12cm, trim = 2cm 2cm 2cm 2cm, clip]{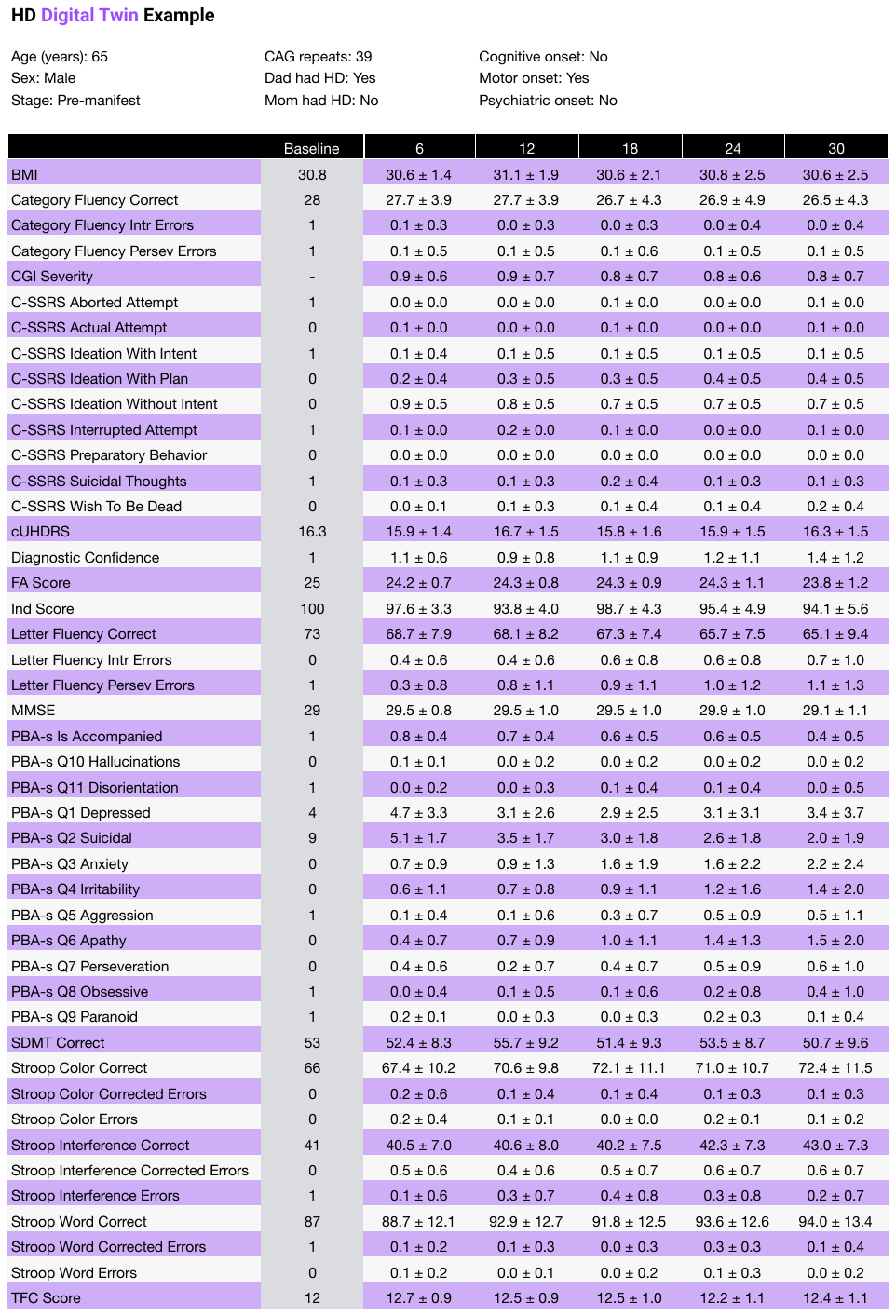}
    \caption{An example twin record for a single HD patient is shown, which reports the mean and standard deviation of the twins for that patient. A sample of static baseline information is given at the top of the record. A sample of longitudinal outcomes predicted by the model are supplied as rows, with columns indicating the time in months since baseline visit. The number of longitudinal records have been truncated.}
    \label{fig:hd-example-twin}
\end{figure}

\begin{figure}
    \centering
    \includegraphics[width=12cm, trim = 2cm 2cm 2cm 2cm, clip]{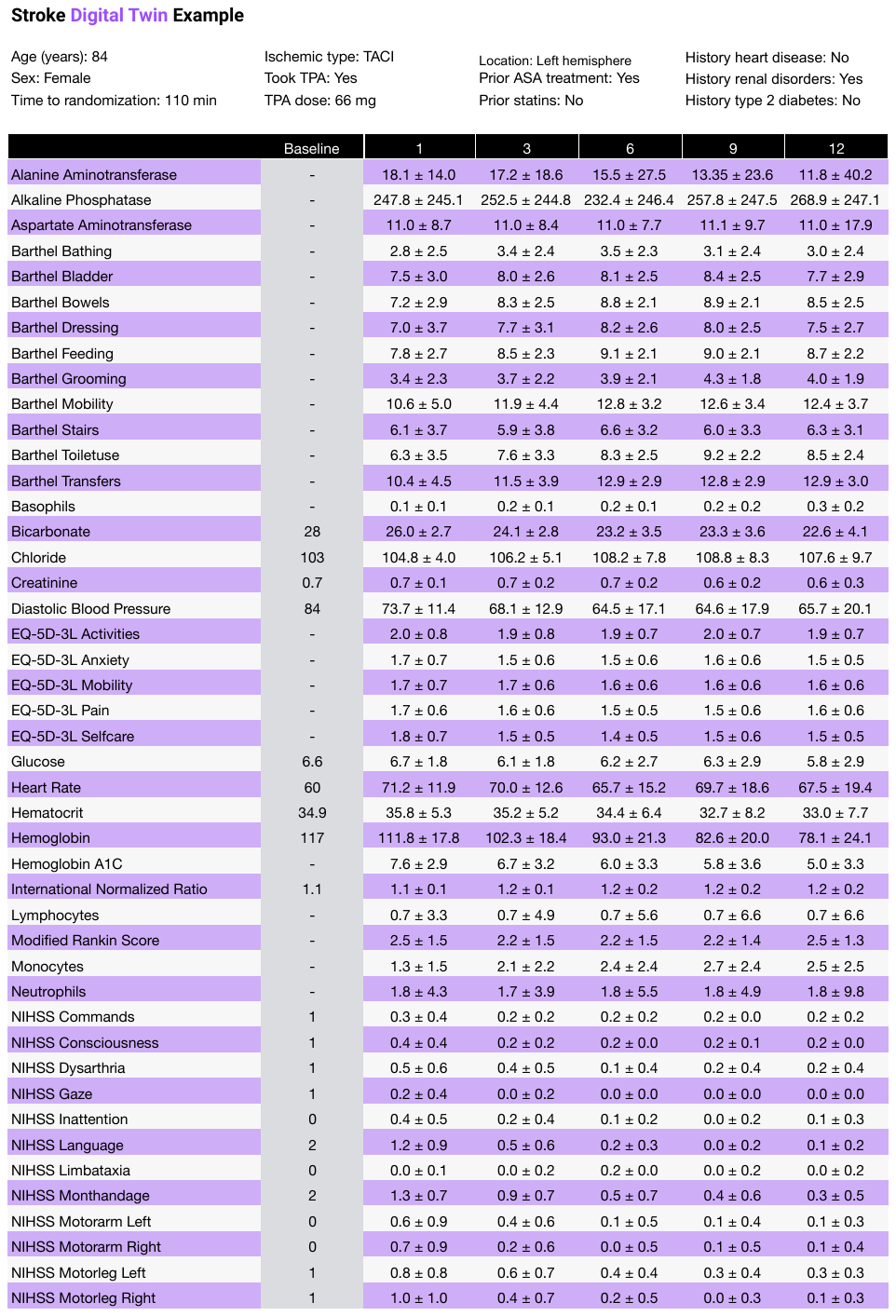}
    \caption{An example twin record for a single Stroke patient is shown, which reports the mean and standard deviation of the twins for that patient. A sample of static baseline information is given at the top of the record. A sample of longitudinal outcomes predicted by the model are supplied as rows, with columns indicating the time in months since baseline visit. The number of longitudinal records have been truncated.}
    \label{fig:str-example-twin}
\end{figure}

\begin{figure}
    \centering
    \includegraphics[width=12cm, trim = 2cm 19cm 2cm 2cm, clip]{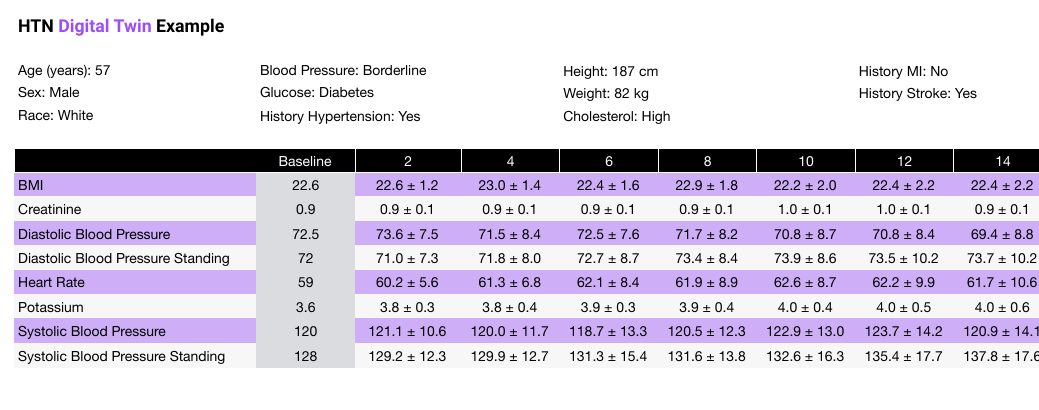}
    \caption{An example twin record for a single HTN patient is shown, which reports the mean and standard deviation of the twins for that patient A sample of static baseline information is given at the top of the record. A sample of longitudinal outcomes predicted by the model are supplied as rows, with columns indicating the time in months since baseline visit.}
    \label{fig:htn-example-twin}
\end{figure}

\begin{figure}
    \centering
    \includegraphics[width=12cm, trim = 2cm 20cm 2cm 2cm, clip]{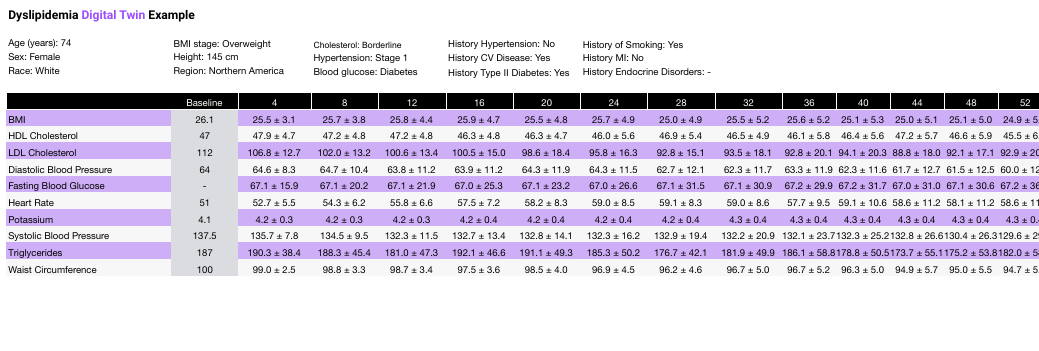}
    \caption{An example twin record for a single DLD patient is shown, which reports the mean and standard deviation of the twins for that patient. A sample of static baseline information is given at the top of the record. A sample of longitudinal outcomes predicted by the model are supplied as rows, with columns indicating the time in months since baseline visit.}
    \label{fig:dld-example-twin}
\end{figure}

\begin{figure}
    \centering
    \includegraphics[width=12cm, trim = 2cm 18cm 2cm 2cm, clip]{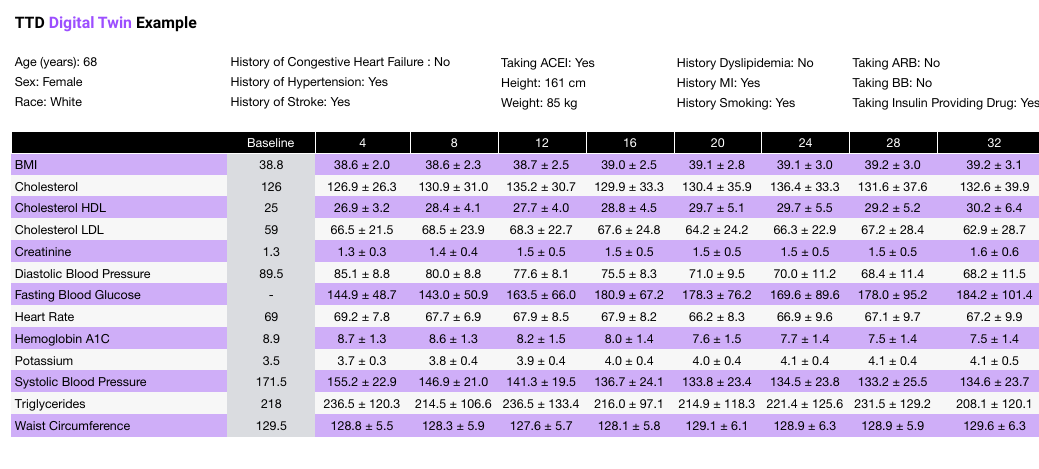}
    \caption{An example twin record for a single TTD patient is shown, which reports the mean and standard deviation of the twins for that patient. A sample of static baseline information is given at the top of the record. A sample of longitudinal outcomes predicted by the model are supplied as rows, with columns indicating the time in months since baseline visit.}
    \label{fig:ttd-example-twin}
\end{figure}

\begin{figure}
    \centering
    \includegraphics[width=12cm, trim = 2cm 17cm 2cm 2cm, clip]{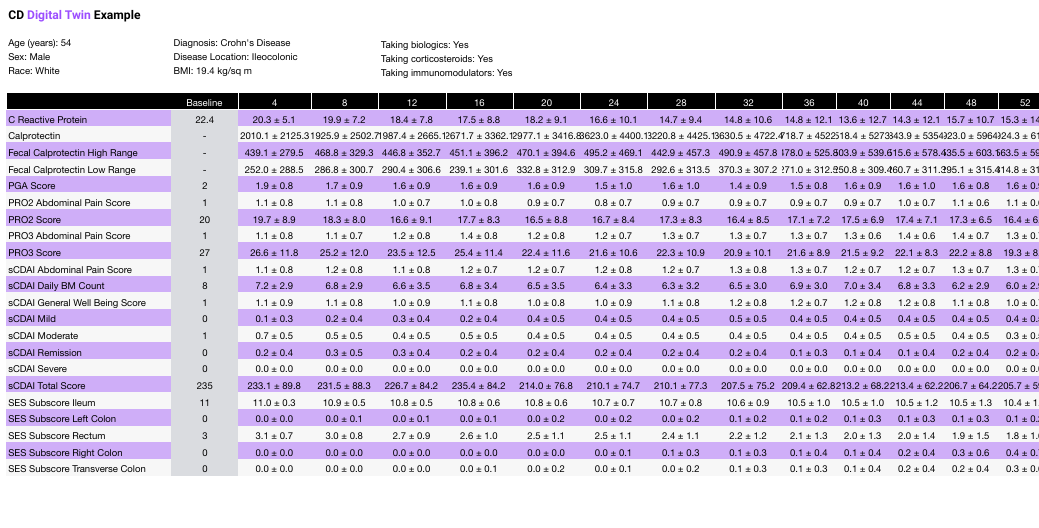}
    \caption{An example twin record for a single CD patient is shown, which reports the mean and standard deviation of the twins for that patient. A sample of static baseline information is given at the top of the record. A sample of longitudinal outcomes predicted by the model are supplied as rows, with columns indicating the time in months since baseline visit.}
    \label{fig:cd-example-twin}
\end{figure}

\begin{figure}
    \centering
    \includegraphics[width=12cm, trim = 2cm 5cm 2cm 2cm, clip]{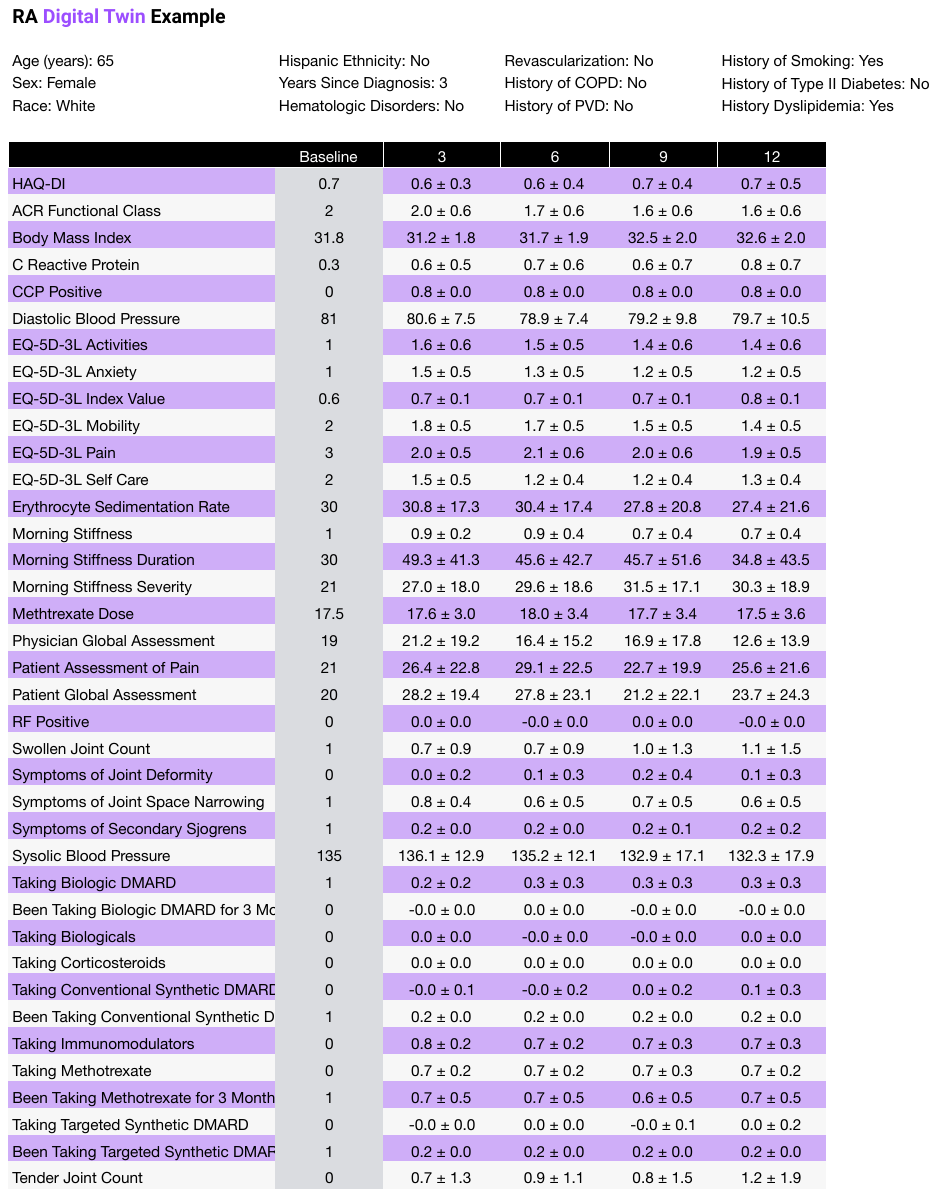}
    \caption{An example twin record for a single RA patient is shown, which reports the mean and standard deviation of the twins for that patient. A sample of static baseline information is given at the top of the record. A sample of longitudinal outcomes predicted by the model are supplied as rows, with columns indicating the time in months since baseline visit.}
    \label{fig:ra-example-twin}
\end{figure}

\begin{figure}
    \centering
    \includegraphics[width=12cm, trim = 2cm 21cm 2cm 2cm, clip]{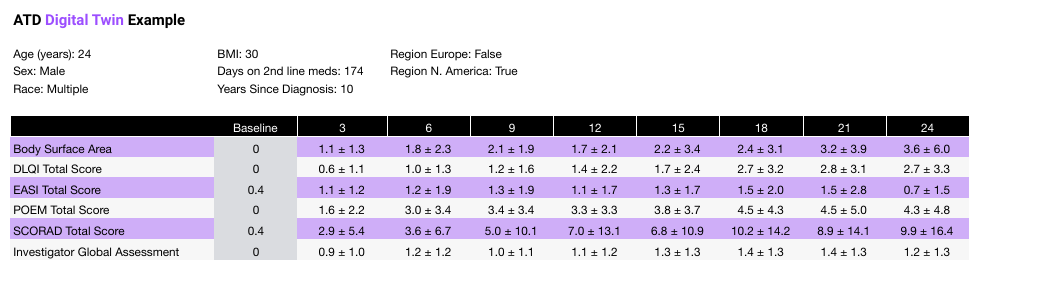}
    \caption{An example twin record for a single ATD patient is shown, which reports the mean and standard deviation of the twins for that patient. A sample of static baseline information is given at the top of the record. A sample of longitudinal outcomes predicted by the model are supplied as rows, with columns indicating the time in months since baseline visit.}
    \label{fig:atd-example-twin}
\end{figure}

\begin{figure}
    \centering
    \includegraphics[width=12cm, trim = 2cm 21cm 2cm 2cm, clip]{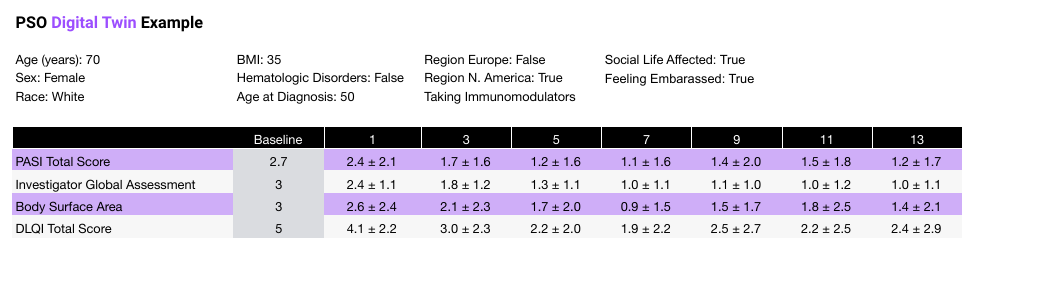}
    \caption{An example twin record for a single PSO patient is shown, which reports the mean and standard deviation of the twins for that patient. A sample of static baseline information is given at the top of the record. A sample of longitudinal outcomes predicted by the model are supplied as rows, with columns indicating the time in months since baseline visit.}
    \label{fig:pso-example-twin}
\end{figure}

Evaluation of the DTG at a per-patient level, rather than at the population level, is more difficult. Since one has only a single observation of a given patient through time, one cannot compare aggregates, expectations, or moments of the twin distribution to the same quantities of the data. A key metric we examine that contains some per-patient level information is the correlation between a patient's trajectory and the expected value derived from their twin. When a measurement of a given patient rises or falls, and the twin also rises and falls in concert, the correlation is high.  To make it easier to visualize what digital twin predictions look like, in Figures~\ref{fig:ad-example-twin} --~\ref{fig:pso-example-twin} we show tables of example digital twins for individual patients in each indication using synthetic baseline data, showing the predicted mean and standard deviation through time for all variables.

Finally, we examine model explainability metrics as a way to assess robustness of our DTGs. We compute feature importance metrics with SHAP to confirm they meet general expectations from disease-specific domain knowledge. Similarly, we asses sensitivity of DTG performance to missing input features, which is a common scenario when trying to use the DTG with clinical data.

\subsection{Mean Predictions for Key Cohorts and Outcomes}

For each indication, we make a DTG conditioned on the multivariate baseline characteristics of each patient. Baseline time refers here to the start of a clinical trial or observational study. Many trajectories of each patient are generated in order to compute expectations or other summary statistics from their digital twins.

A patient's digital twin produces a multivariate probability distribution for all outcomes at all future times, conditioned on data collected from that patient at previous times. One set of useful statistics for summarizing this distribution is the expected values of important clinical outcomes at key future time points. These expected values are essentially point predictions for those future outcomes in the sense that they minimize the mean-squared error under the model distribution. From the law of total expectation, the
predicted mean for a given variable $y$ taken over a population of patients, at a given time $t$, can be written as 
\begin{equation}
    \mu_{\text{pred}}[y(t)] = \mathbb{E}[\mathbb{E}[y_{\text{pred}}(t)|{\bf x}]], 
\end{equation}
where ${\bf x}$ indicates the baseline characteristics of a given patient and
$y_{\text{pred}}(t)| {\bf x}$ a is a sample for $y(t)$ obtained from a patient's digital twin. Thus, the inner expectation $\mathbb{E}[y_{\text{pred}}(t)|{\bf x}]$ is the predicted mean for the variable obtained from a patient's digital twin, and the outer expectation takes the average of the population.  

\begin{figure}
    \centering
    \includegraphics[width=11cm, clip]{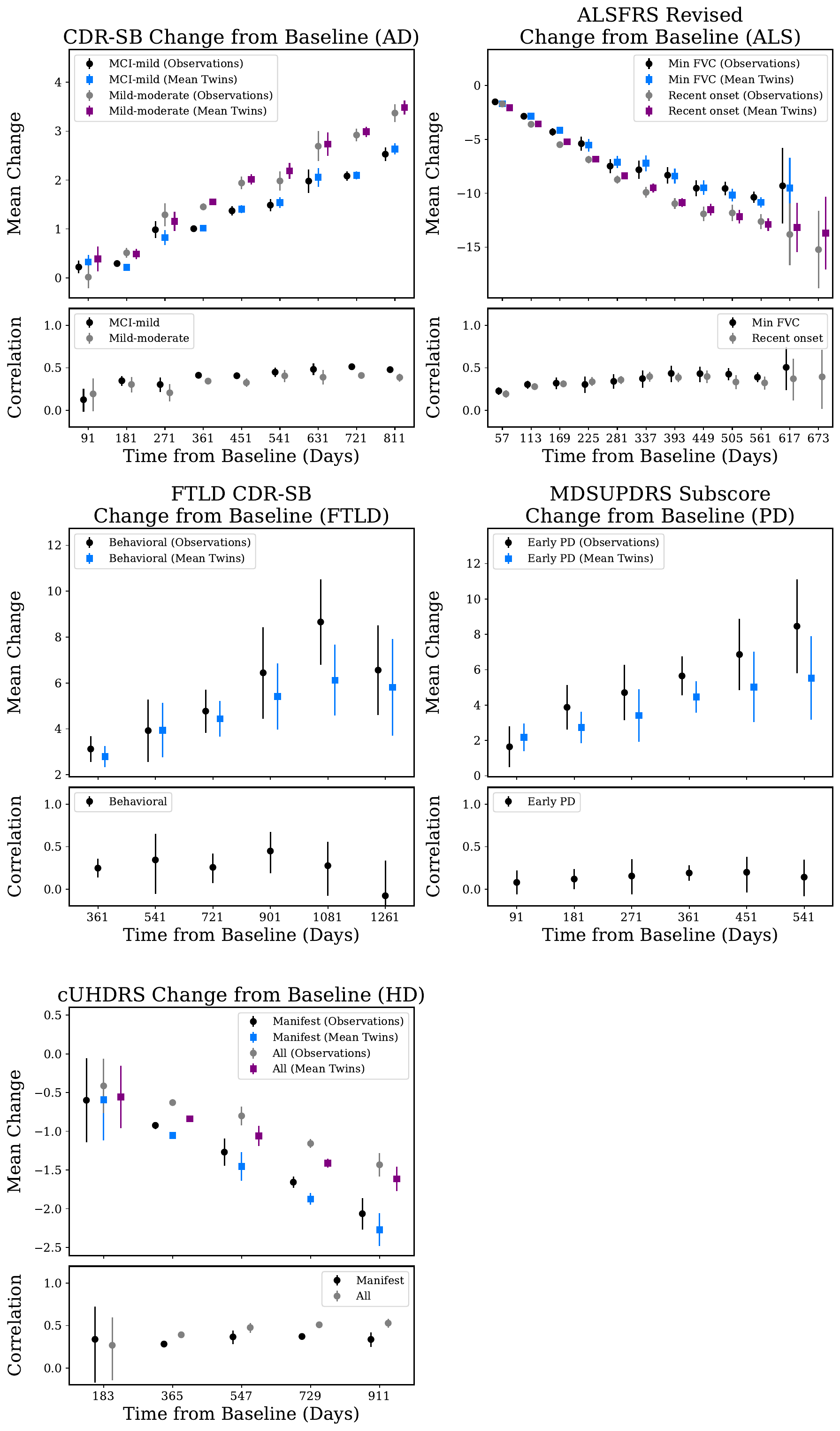}
    \caption{Observed and predicted progression of key outcomes for indications in the Neurodegenerative therapeutic area. In each plot, we selected a key outcome and up to two cohorts (see \Tabs{outcomes-neuro}{cohort-def-neuro} for descriptions of outcomes and cohorts, respectively). The {\bf top panel} compares the observed and predicted mean progression. The {\bf bottom panel} reports Pearson correlation between observations and mean twins over time as a measure of discriminative performance of the model. Error bars represent 95\% confidence intervals. DTGs fit well mean progression across indications and clinically relevant cohorts.} 
    \label{fig:progression_neuro}
\end{figure}

\begin{figure}
    \centering
    \includegraphics[width=11cm, clip]{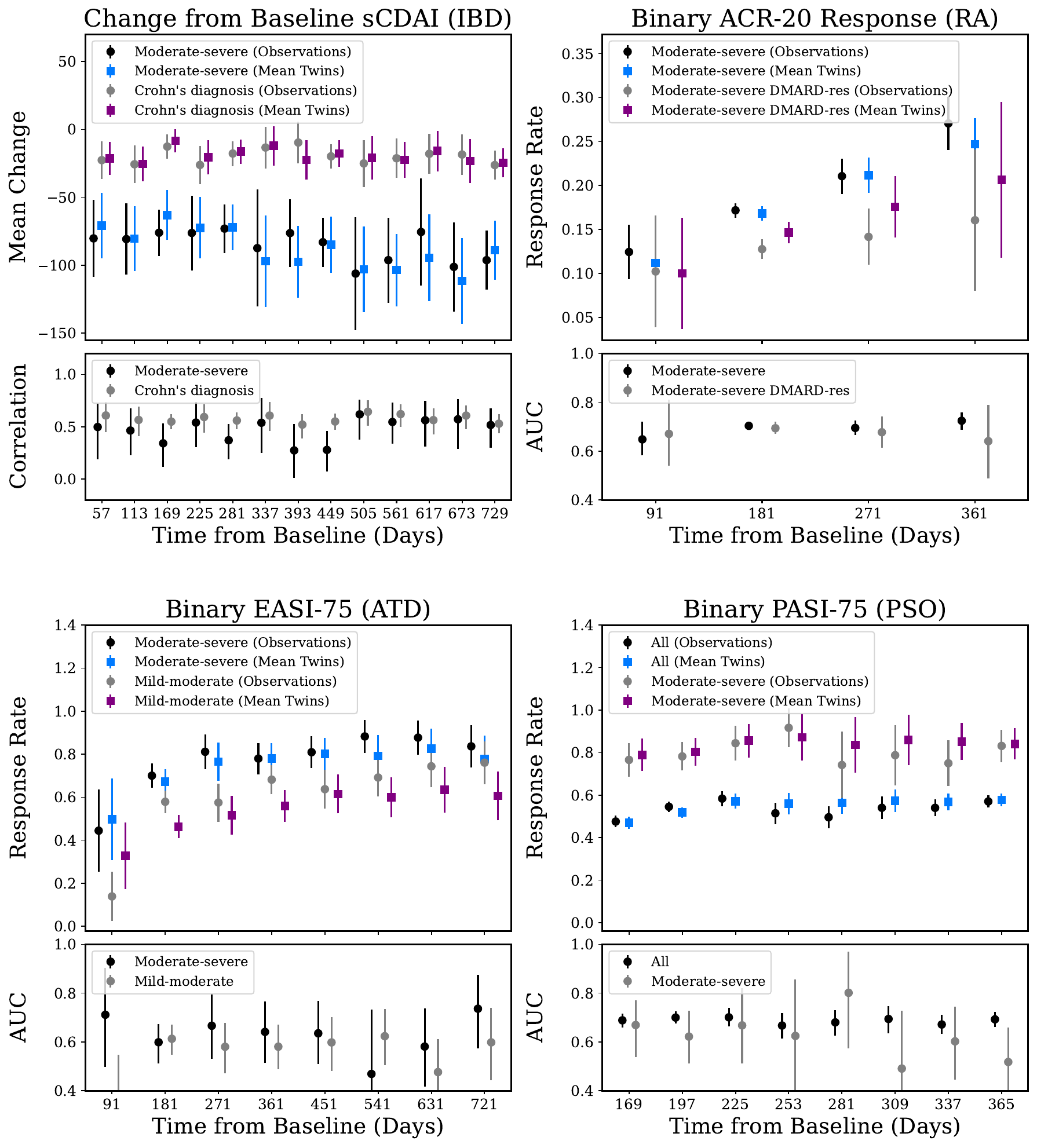}
    \caption{Observed and predicted mean progression of key outcomes for indications in the Immunology \& Inflammation therapeutic area. In each plot, we selected a key outcome and up to two cohorts (see \Tabs{outcomes-ini}{cohort-def-ini} for descriptions of outcomes and cohorts, respectively). The {\bf top panel} compares the observed and predicted mean progression. For binary outcomes, we report response rate. The {\bf bottom panel} reports Pearson correlation between observations and mean twins over time for continuous outcomes and AUC for binary outcomes, both as a measure of discriminative performance of the model. Error bars represent 95\% confidence intervals. DTGs fit well mean progression across indications and clinically relevant cohorts.}
    \label{fig:progression_immuno}
\end{figure}

\begin{figure}
    \centering
    \includegraphics[width=11cm, clip]{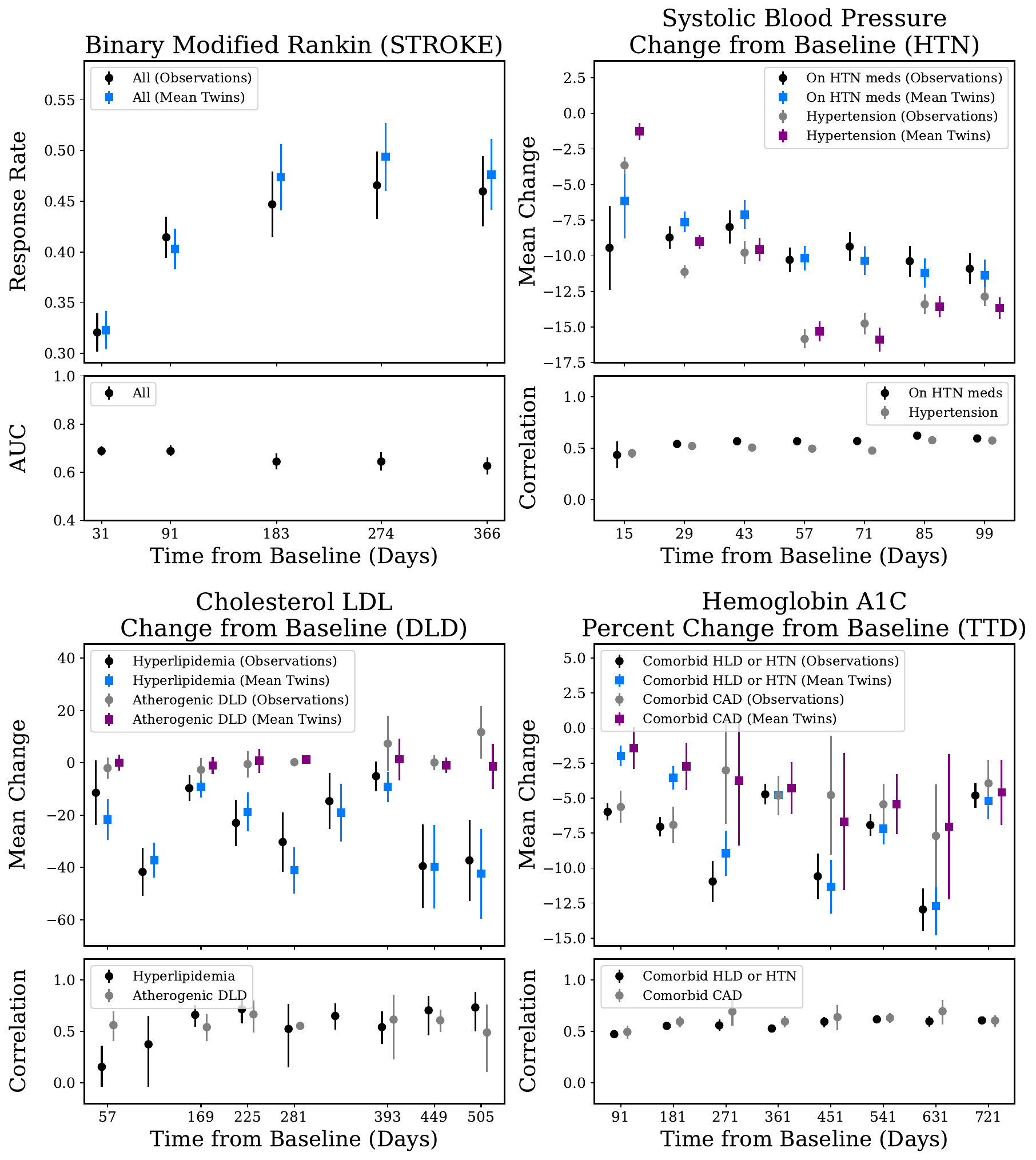}
    \caption{Observed and predicted mean progression of key outcomes for indications in the General Medicine therapeutic area. In each plot, we selected a key outcome and up to two cohorts (see \Tabs{outcomes-gm}{cohort-def-gm} for descriptions of outcomes and cohorts, respectively). The {\bf top panel} compares the observed and predicted mean progression. For binary outcomes, we report response rate. The {\bf bottom panel} reports Pearson correlation between observations and mean twins over time for continuous outcomes and AUC for binary outcomes, both as a measure of discriminative performance of the model. Error bars represent 95\% confidence intervals. DTGs fit well mean progression across indications and clinically relevant cohorts.}
    \label{fig:progression_general}
\end{figure}

In ~\Fig{progression_neuro},~\Fig{progression_immuno}, and~\Fig{progression_general}, we consider a key outcome for each indication and report mean progression over time, comparing data and DTG predictions at the population level. When available, we report progression for two clinically relevant cohorts. For binary outcomes, we compare predicted and observed response rates over time. We also report Pearson correlation between observations and DTG predictions for continuous outcomes and AUC for binary outcomes. Correlations provide an important measure of goodness-of-fit beyond the population-level mean trajectories that measure the strength of the linear relationship between observed and predicted changes in clinical variables at the patient level. Similarly, AUC measures discriminative performance for binary outcomes.

These figures demonstrate that the DTGs are effective at generating digital twins with low bias and good discrimination.  The predicted progression from the digital twins tracks the observed outcomes well, and distinct cohorts show distinct predicted progressions that match observations well.  The degree to which the digital twins capture changes in progression across populations is quite stark and suggests that the model is doing a good job at utilizing prognostic information contained in the data as well as modeling the time dynamics of the disease across all indications.  The digital twins have good correlation with the observations, ranging between 0.3 and 0.5 across nearly all Neurodegeneration indications, and between 0.5 and 0.7 for Immunology \& Inflammation as well as General Medicine.  Note that, to simplify presentation, these figures only show a single outcome for each indication; the DTGs predict multiple outcomes that are common clinical trial endpoints for each indication.

\begin{SCfigure}[][ht]
    \centering
    \includegraphics[width=8cm, trim = 0cm 5cm 0cm 7cm, clip]{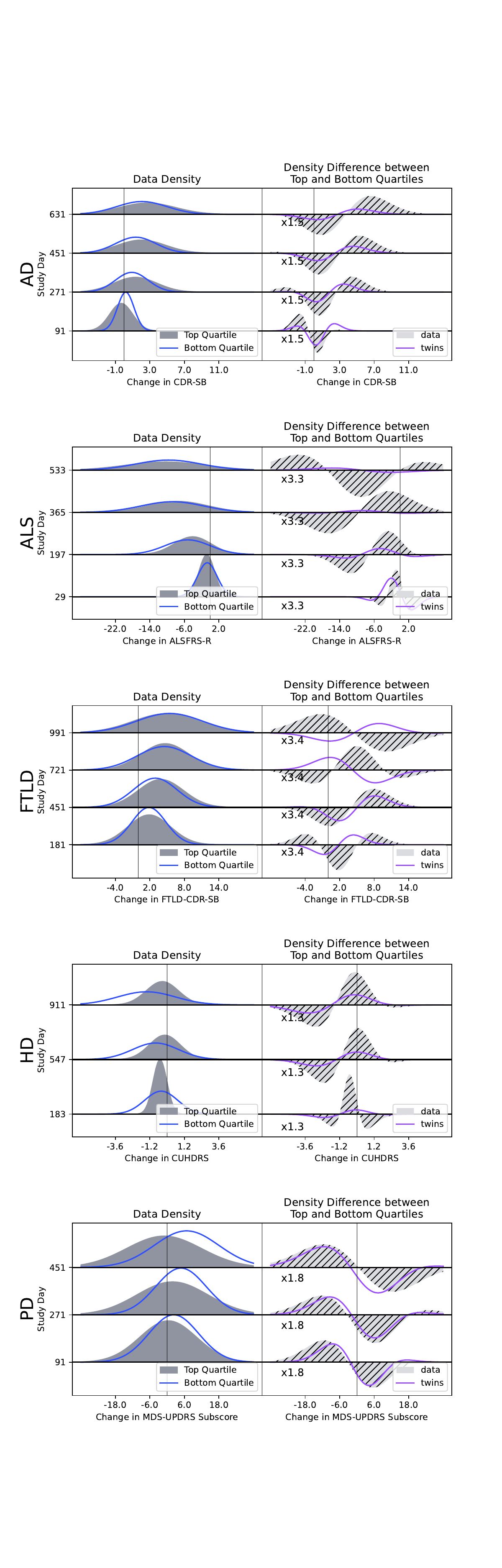}
    \caption{We examine the ability of each DTG per indication in the Neurodegeneration therapeutic area to accurately capture distribution changes of the data for different strata of the overall training population. Each plot stratifies the held-out data into quartiles according to the baseline score indicated in the x-axis label. In the left column, the density of the top and bottom quartiles of the data are depicted as a Gaussian density with mean and standard deviation obtained from the data. The difference between the these densities is depicted in the right column as a solid hashed difference density. For small differences we magnify the density difference by a factor noted in text on the plot. The difference between the twin distribution conditioned on the top quartile and the twin distribution conditioned on the bottom quartiles is overlaid in the right column.}
    \label{fig:difference_density_neuro}
\end{SCfigure}

\begin{SCfigure}[][ht]
    \centering
    \includegraphics[width=8cm, trim = 0cm 5cm 0cm 5cm, clip]{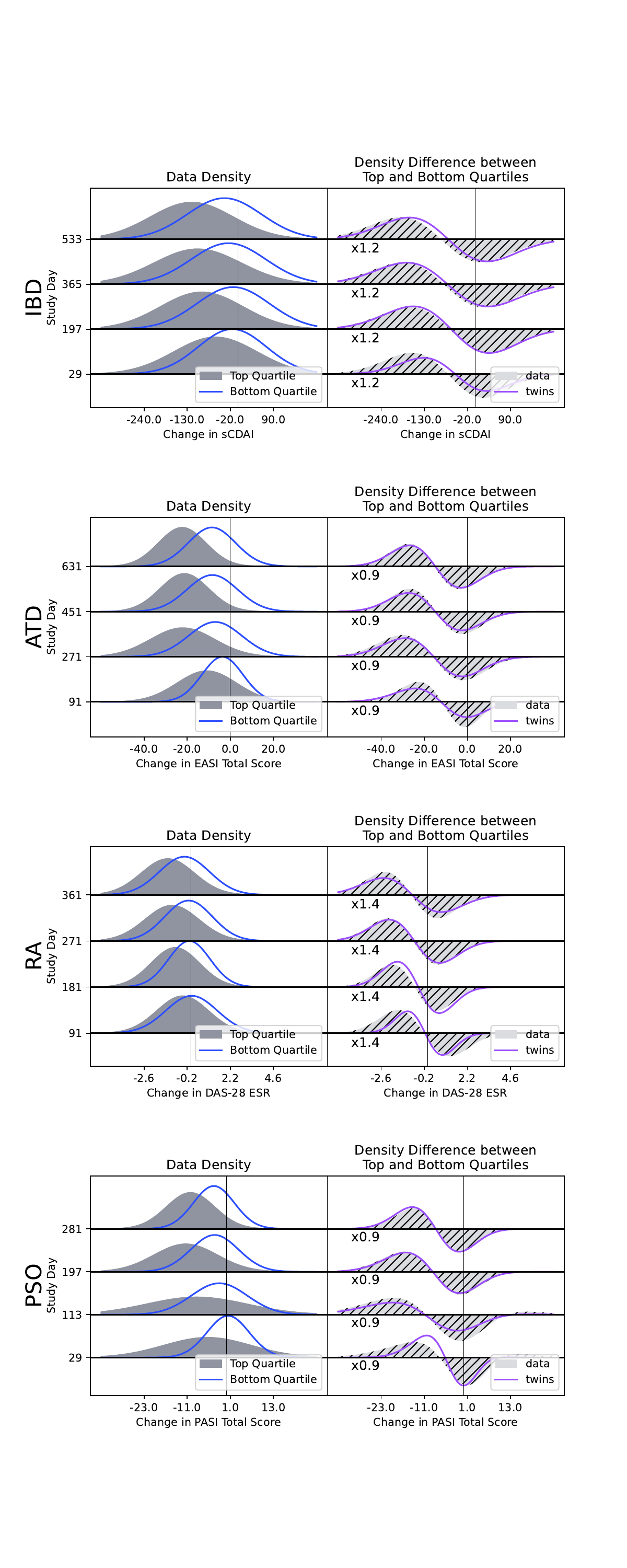}
    \caption{We examine the ability of each DTG per indication in the Immunology \& Inflammation therapeutic area to accurately capture distribution changes of the data for different strata of the overall training population. Each plot stratifies the held-out data into quartiles according to the baseline score indicated in the x-axis label. In the left column, the density of the top and bottom quartiles of the data are depicted as a Gaussian density with mean and standard deviation obtained from the data. The difference between the these densities is depicted in the right column as a solid hashed difference density. For small differences we magnify the density difference by a factor noted in text on the plot. The difference between the twin distribution conditioned on the top quartile and the twin distribution conditioned on the bottom quartiles is overlaid in the right column.}
    \label{fig:difference_density_ini}
\end{SCfigure}

\begin{SCfigure}[][ht]
    \centering
    \includegraphics[width=8cm, trim = 0cm 5cm 0cm 5cm, clip]{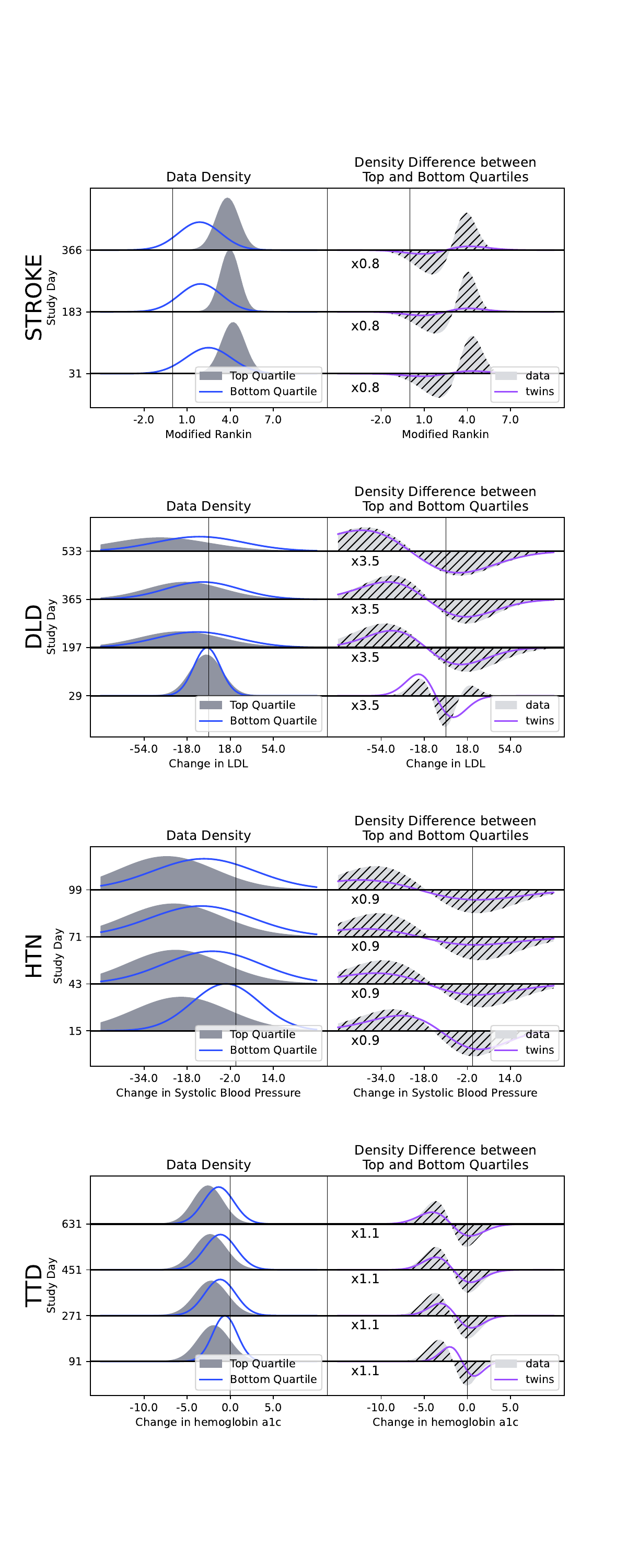}
    \caption{We examine the ability of each DTG per indication in the General Medicine therapeutic area to accurately capture distribution changes of the data for different strata of the overall training population. Each plot stratifies the held-out data into quartiles according to the baseline score indicated in the x-axis label. Note that in the case of Stroke, Modified Rankin is conditioned on the value at 91 days, rather than at baseline, as this feature is not measured at baseline. In the left column, the density of the top and bottom quartiles of the data are depicted as a Gaussian density with mean and standard deviation obtained from the data. The difference between the these densities is depicted in the right column as a solid hashed difference density. For small differences we magnify the density difference by a factor noted in text on the plot. The difference between the twin distribution conditioned on the top quartile and the twin distribution conditioned on the bottom quartiles is overlaid in the right column.}
    \label{fig:difference_density_gm}
\end{SCfigure}

In ~\Fig{difference_density_neuro}, ~\Fig{difference_density_ini},~\Fig{difference_density_gm} we focus on how well the DTGs in each indication are able to model a conditional distribution. Rather than using clinical cohorts, we stratify the overall training population into quartiles, select the top and bottom quartiles, and examine how well the model can capture the mean and standard deviation of these different sub-populations. The difference between the population densities of the different sub-populations of both the digital twins and data are compared in the right column of these figures. The data densities of the top and bottom quartiles are shown in the left column. Generally, we see that the DTG is able to not only well approximate the density of the data, but also the change in density of the data from one sub-population to the next. There are times when the twin density difference disagrees with the data, most often when the difference between quartiles is small.

\subsection{Comprehensive Evaluation}

\begin{figure}
    \centering
    \includegraphics[width=0.8\textwidth]{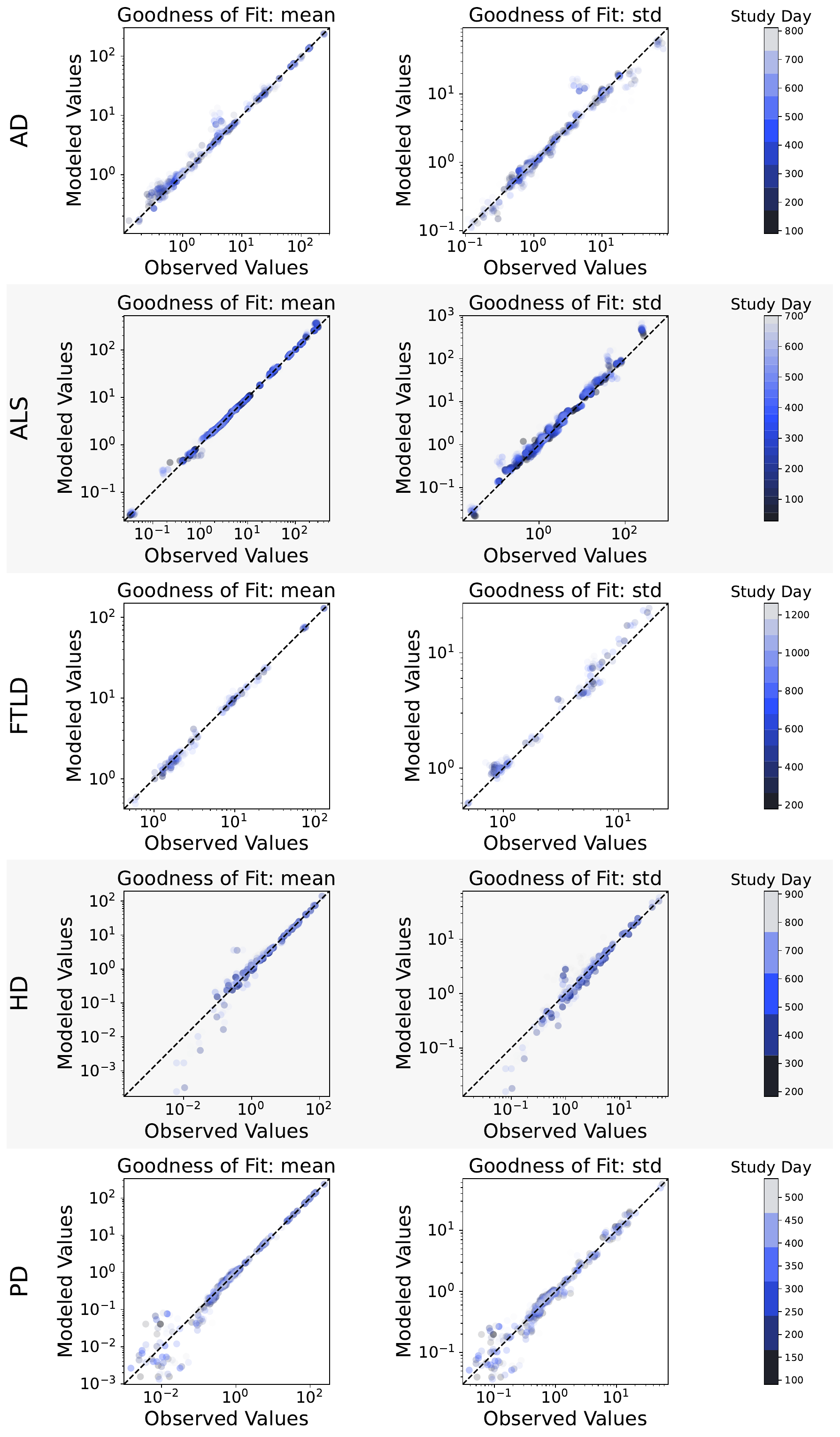}
    \caption{Goodness of fit for mean (left column) and standard deviation (right column) in the Neurodegeneration therapeutic area, with the indication for a given plot indicated by abbreviation on the left axis. All plots report on the principle cohort identified in \Tab{cohort-def-neuro}. The mean and standard deviation for all change from baseline longitudinal outcomes are evaluated at discrete time bins indicated by the color bar for both the data and the digital twins. As the twins should represent the data well, we ideally expect the markers to for either moment to all lie on the dotted diagonal line.}
    \label{fig:gof_neuro}
\end{figure}

\begin{figure}
    \centering
    \includegraphics[width=0.9\textwidth]{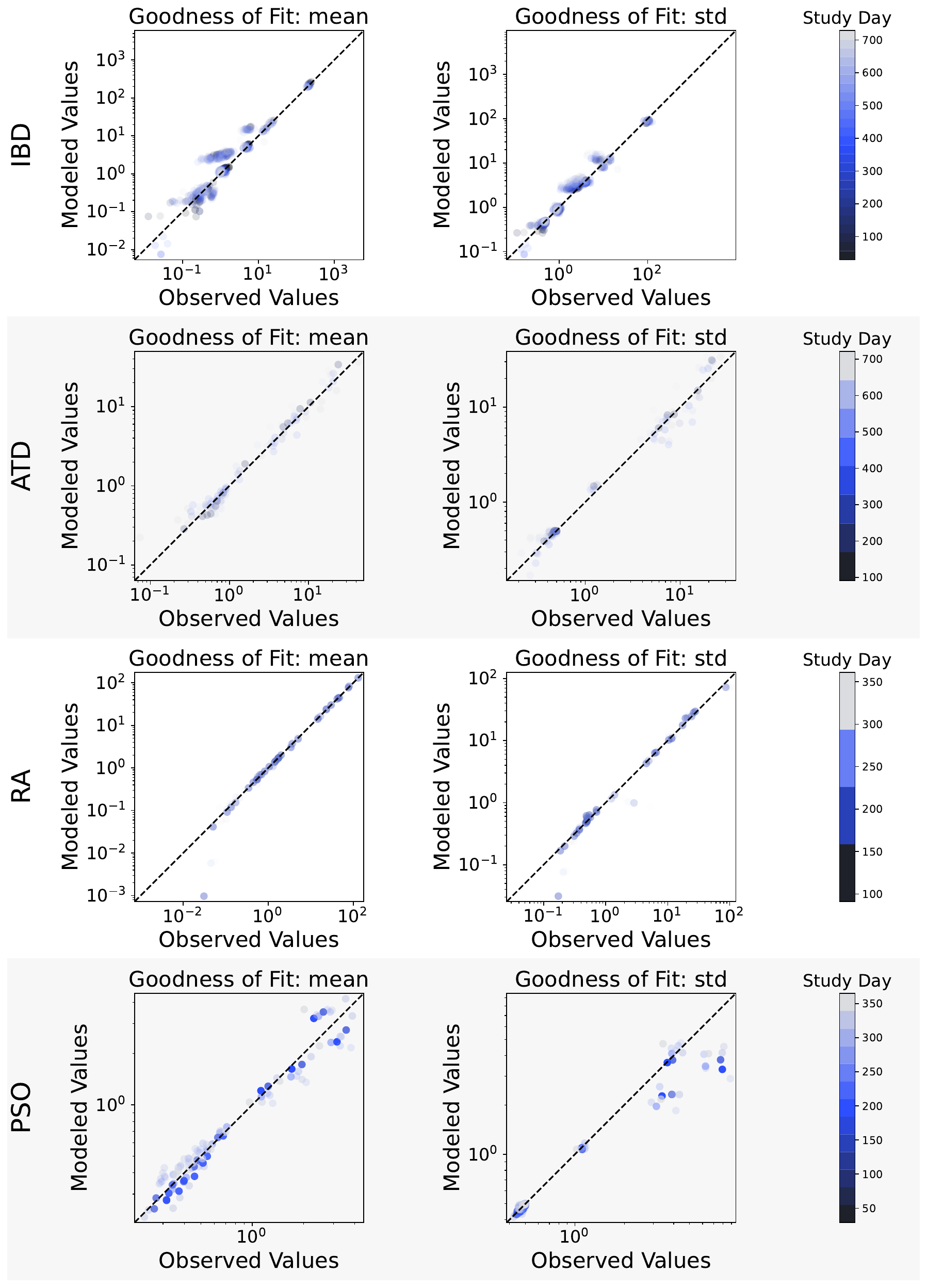}
    \caption{Goodness of fit for mean (left columns) and standard deviation (right column) in the Immunology and Inflammation therapeutic area, with the indication for a given plot indicated by abbreviation on the left axis. All plots report on the principle cohort identified in \Tab{cohort-def-ini}. The mean and standard deviation for all change from baseline longitudinal outcomes are evaluated at discrete time bins indicated by the color bar for both the data and the digital twins. As the twins should represent the data well, we ideally expect the markers to for either moment to all lie on the dotted diagonal line.}
    \label{fig:gof_ini}
\end{figure}

\begin{figure}
    \centering
    \includegraphics[width=0.9\textwidth]{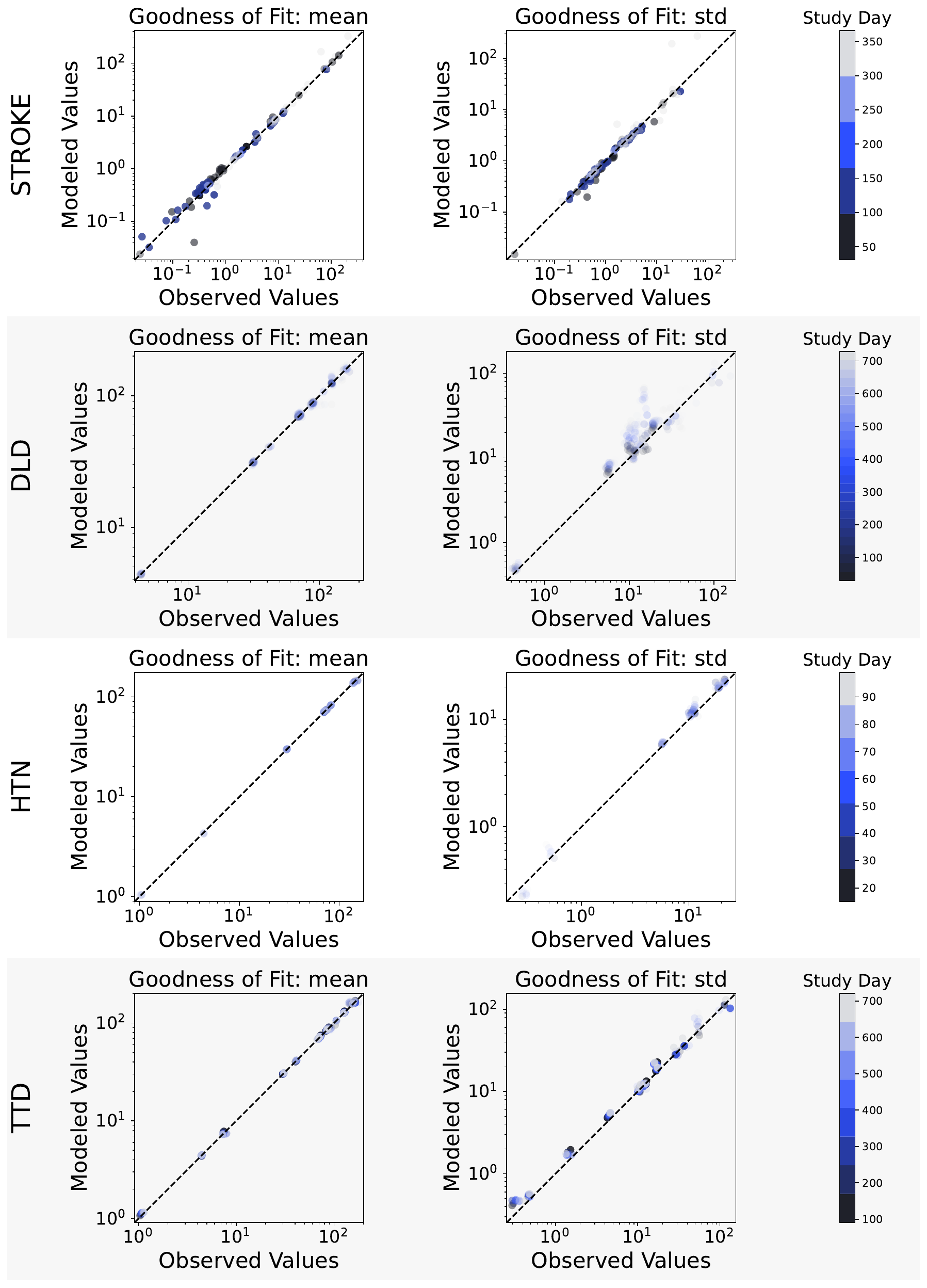}
    \caption{Goodness of fit for mean (left column) and standard deviation (right column) in the General Medicine therapeutic area, with the indication for a given plot indicated by abbreviation on the left axis. All plots report on the principle cohort identified in \Tab{cohort-def-gm}. The mean and standard deviation for all change from baseline longitudinal outcomes are evaluated at discrete time bins indicated by the color bar for both the data and the digital twins. As the twins should represent the data well, we ideally expect the markers to for either moment to all lie on the dotted diagonal line.}
    \label{fig:gof_gm}
\end{figure}

Metrics and figures discussed above focused on a few key clinical outcomes in each indication, but each of these diseases is multifaceted, and digital twins of patients in each indication need to model many other longitudinal variables including laboratory tests, biomarkers, vitals, and additional clinical assessments. In the left panel of \Fig{gof_neuro}, \Fig{gof_ini}, and \Fig{gof_gm}, for each indication, we compare predicted and
observed population means for all variables and times. A well-calibrated model should correctly reproduce the means, so that data points accumulate along the diagonal. Marker opacity is scaled according to the number of observations available for that particular outcome and time, and color indicates time since baseline. A perfectly calibrated model would have all predicted means fall along a line of unit slope and zero intercept. 

So far, we have focused on evaluating the mean twin predictions. DTGs,
however, characterize a full multivariate distribution over time conditioned on a patient's baseline characteristics. The right panel of \Fig{gof_neuro}, \Fig{gof_ini}, and \Fig{gof_gm} compares standard
deviations. For each variable and time, the x-axis reports the observed
population standard deviation and y-axis the corresponding prediction. Marker
opacity and colors represent size of the population and time, as for plots comparing the means. Note that from the law of total variance, the predicted population standard deviation for a given variable $y$, at a given time $t$, can be written as 
\begin{equation}
    \sigma_{\text{pred}}[y(t)] = \left(\text{Var}[\mathbb{E}[y_{\text{pred}}(t)|{\bf x}]] + \mathbb{E}[\text{Var}[y_{\text{pred}}(t)|{\bf x}]]\right)^{1/2},
    \label{eq:marginal_var}
\end{equation}
where ${\bf x}$ indicates the baseline characteristics of a given patient and
$y_{\text{pred}}(t)| {\bf x}$ a is a sample for $y(t)$ obtained from a patient's digital twin. The predicted variance has two contributions:
the first is the population variance of the per-patient mean twins
$\mathbb{E}[y_{\text{pred}}(t)|{\bf x}]$, and the second is the population mean of the per-patient variances $\text{Var}[y_{\text{pred}}(t)|{\bf x}]$. 

We see good agreement between the predicted and observed values for these first and second moments across therapeutic areas and indications.  Outliers are rare, and tend to coincide with smaller sample sizes for evaluation.  This gives confidence that the DTGs are correctly capturing behavior of hte distributions across time for each individual variable.

\begin{figure}
    \centering
    \includegraphics[width=\textwidth]{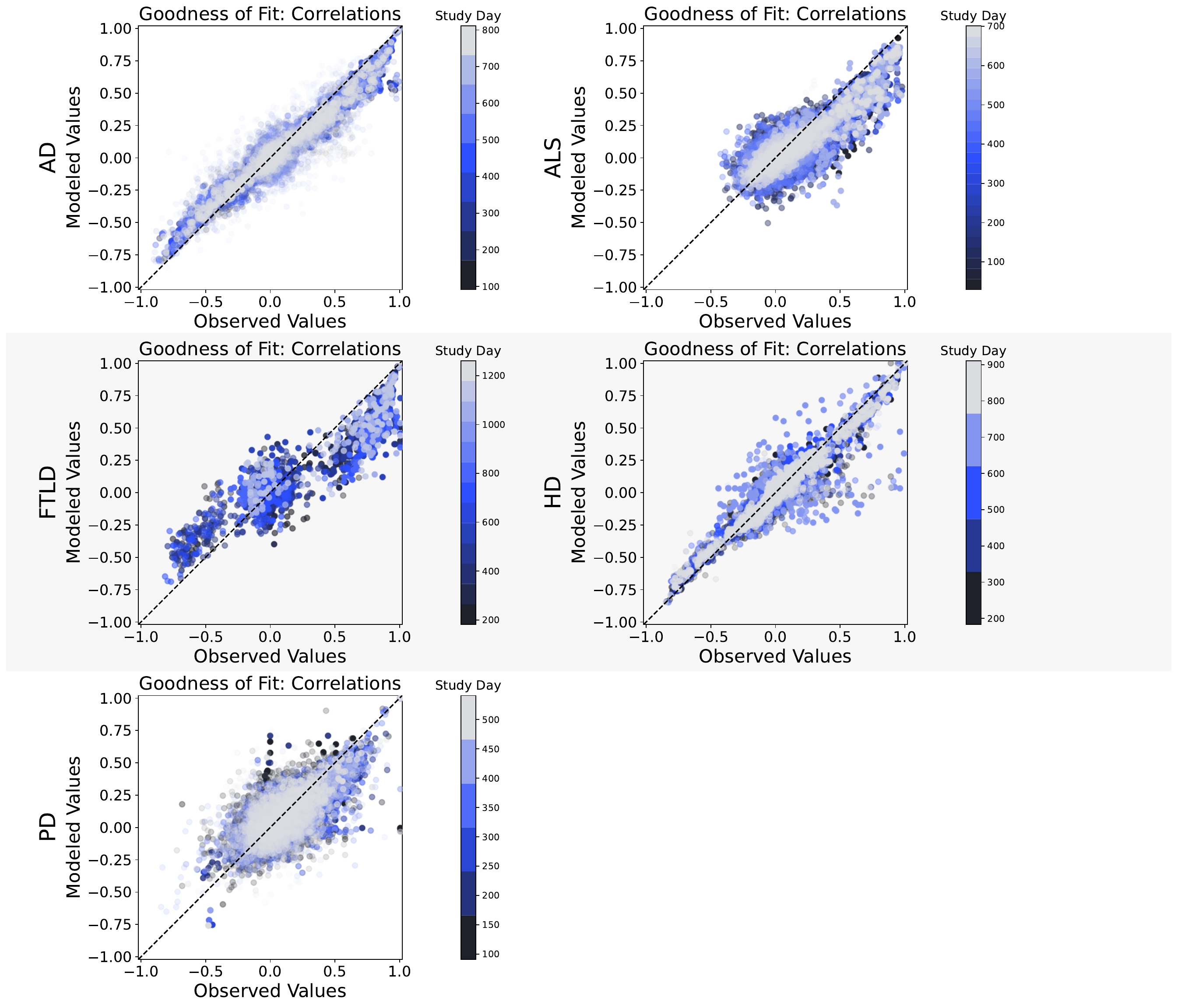}
    \caption{Goodness of fit for cross correlations in the Neurodegeneration therapeutic area, with the indication for a given plot indicated by abbreviation on the left axis. All plots report on the principle cohort identified in \Tab{cohort-def-neuro}. Cross-correlations between all change from baseline longitudinal outcomes are evaluated at discrete time bins indicated by the color bar for both the data and the digital twins. As the twins should represent the data well, we ideally expect these cross-correlations to all lie on the dotted diagonal line.}
    \label{fig:galaxy_neuro}
\end{figure}

\begin{figure}
    \centering
    \includegraphics[width=\textwidth]{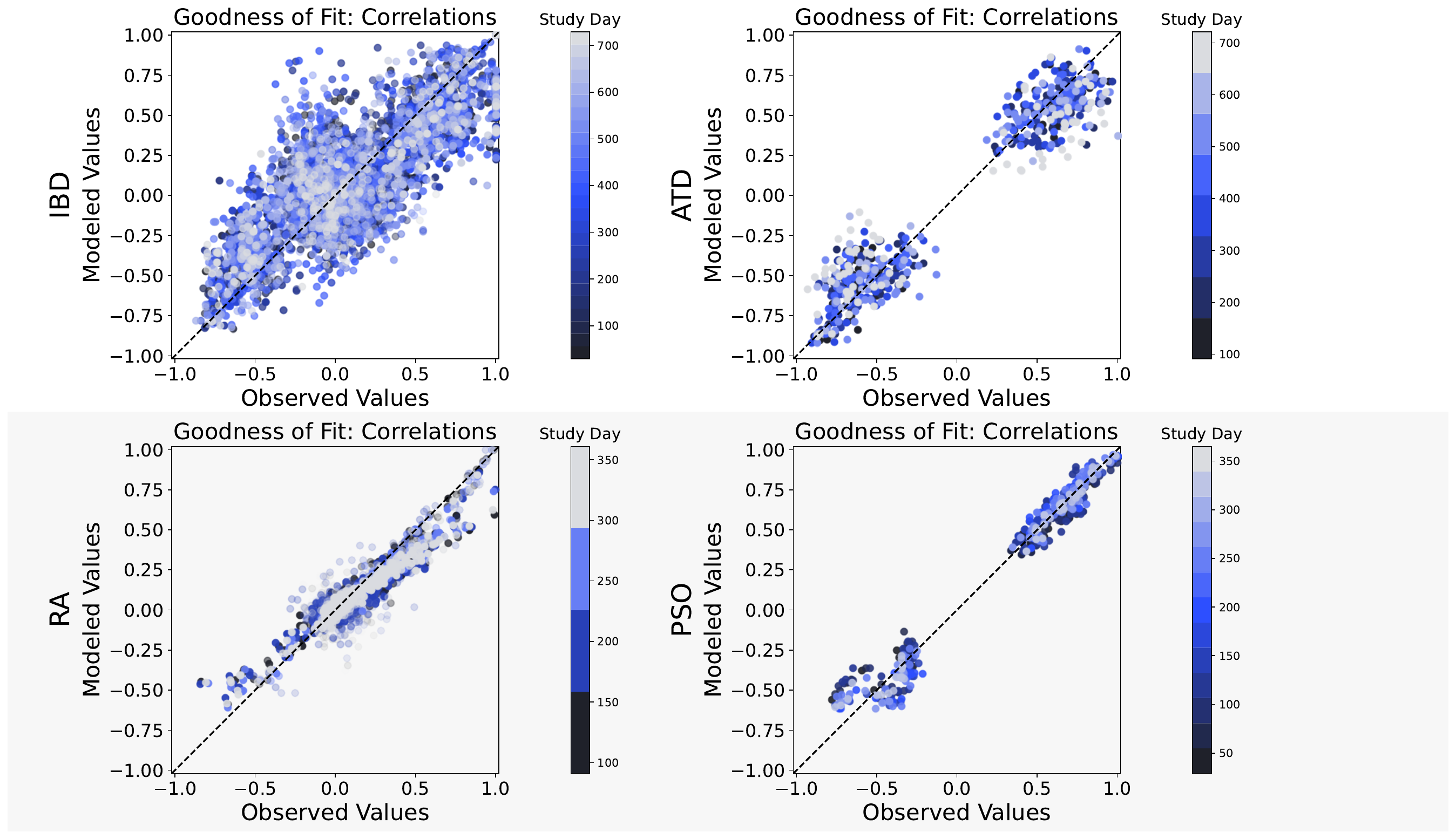}
    \caption{Goodness of fit for cross correlations in the Immunology and Inflammation therapeutic area, with the indication for a given plot indicated by abbreviation on the left axis. All plots report on the principle cohort identified in \Tab{cohort-def-ini}. Cross-correlations between all change from baseline longitudinal outcomes are evaluated at discrete time bins indicated by the color bar for both the data and the digital twins. As the twins should represent the data well, we ideally expect these cross-correlations to all lie on the dotted diagonal line.}
    \label{fig:galaxy_ini}
\end{figure}

\begin{figure}
    \centering
    \includegraphics[width=\textwidth]{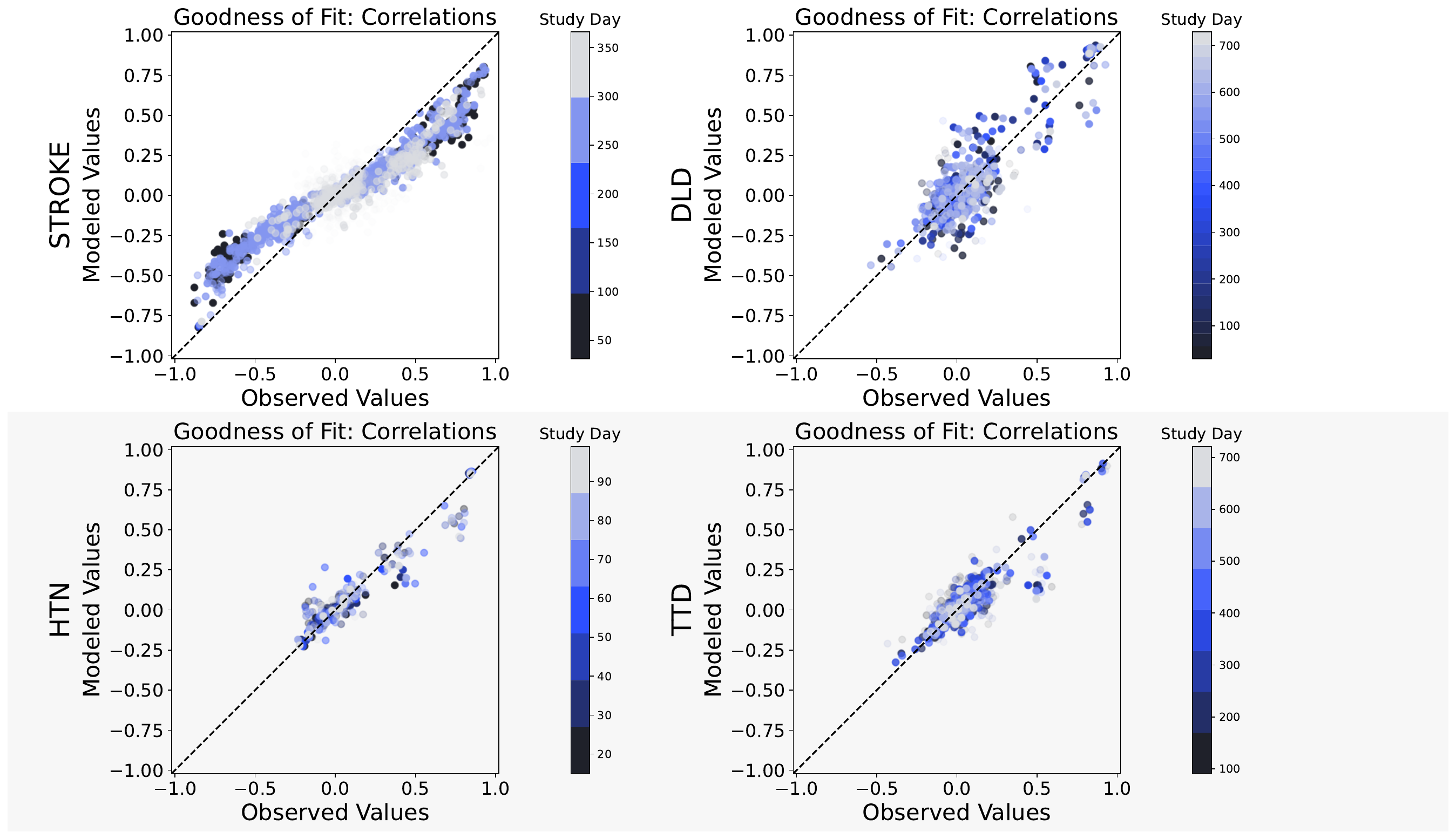}
    \caption{Goodness of fit for cross correlations in the General Medicine therapeutic area, with the indication for a given plot indicated by abbreviation on the left axis. All plots report on the principle cohort identified in \Tab{cohort-def-gm}. Cross-correlations between all change from baseline longitudinal outcomes are evaluated at discrete time bins indicated by the color bar for both the data and the digital twins. As the twins should represent the data well, we ideally expect these cross-correlations to all lie on the dotted diagonal line.}
    \label{fig:galaxy_gm}
\end{figure}

Next, in \Fig{galaxy_neuro}, \Fig{galaxy_ini}, and \Fig{galaxy_gm} we compare observed and predicted equal-time cross-correlations across
variables and times. This allows us to evaluate the full covariance structure of  the digital twins and evaluate the relationships between variables. As in the previous figure, for each pair of variables and for a given time, the x-axis reports the observed cross-correlation and y-axis the corresponding predicted cross-correlation. Marker opacity and colors represent the size of the population and time, as for the other
plots. As for the variances, the predicted correlations receive contributions
both from the correlation across the population of the mean twin predictions, as well as from the correlations of the twins for each patient. From the law of total covariance, the predicted correlation between variables $y$ and $z$ at time $t$ can be written as
\begin{equation}
    \rho_{\text{pred}}[y(t), z(t)] = \frac{\text{Cov}[\mathbb{E}[y_{\text{pred}}(t)|{\bf x}], \mathbb{E}[z_{\text{pred}}(t)|{\bf x}]] + \mathbb{E}[\text{Cov}[y_{\text{pred}}(t), z_{\text{pred}}(t)|{\bf x}]]}{\sigma_{\text{pred}}[y(t)]\sigma_{\text{pred}}[z(t)]},
\end{equation}
where ${\bf x}$ indicates the baseline characteristics of a given patient,
$y_{\text{pred}}(t)| {\bf x}$ and $z_{\text{pred}}(t)| {\bf x}$ 
are samples for $y(t)$ and $z(t)$ obtained from a patient's digital twin. The standard deviations in the denominator are defined in
\Eq{marginal_var}. As for the variances, predicted correlations also receive two contributions. The first term in the numerator is the population covariance of the mean twins and the second term is the population mean of the covariance for each patient. 

We can see that the correlations are generally well modeled, with the areas of largest deviation at small values of correlations.  Interestingly, one can look at the size of correlations across indications and therapeutic areas to gauge the relationships in variables per indication.

\subsection{Model Explainability}

\begin{figure}
    \centering
    \includegraphics[width=14cm, trim = 3.75cm 0cm 2cm 0cm, clip]{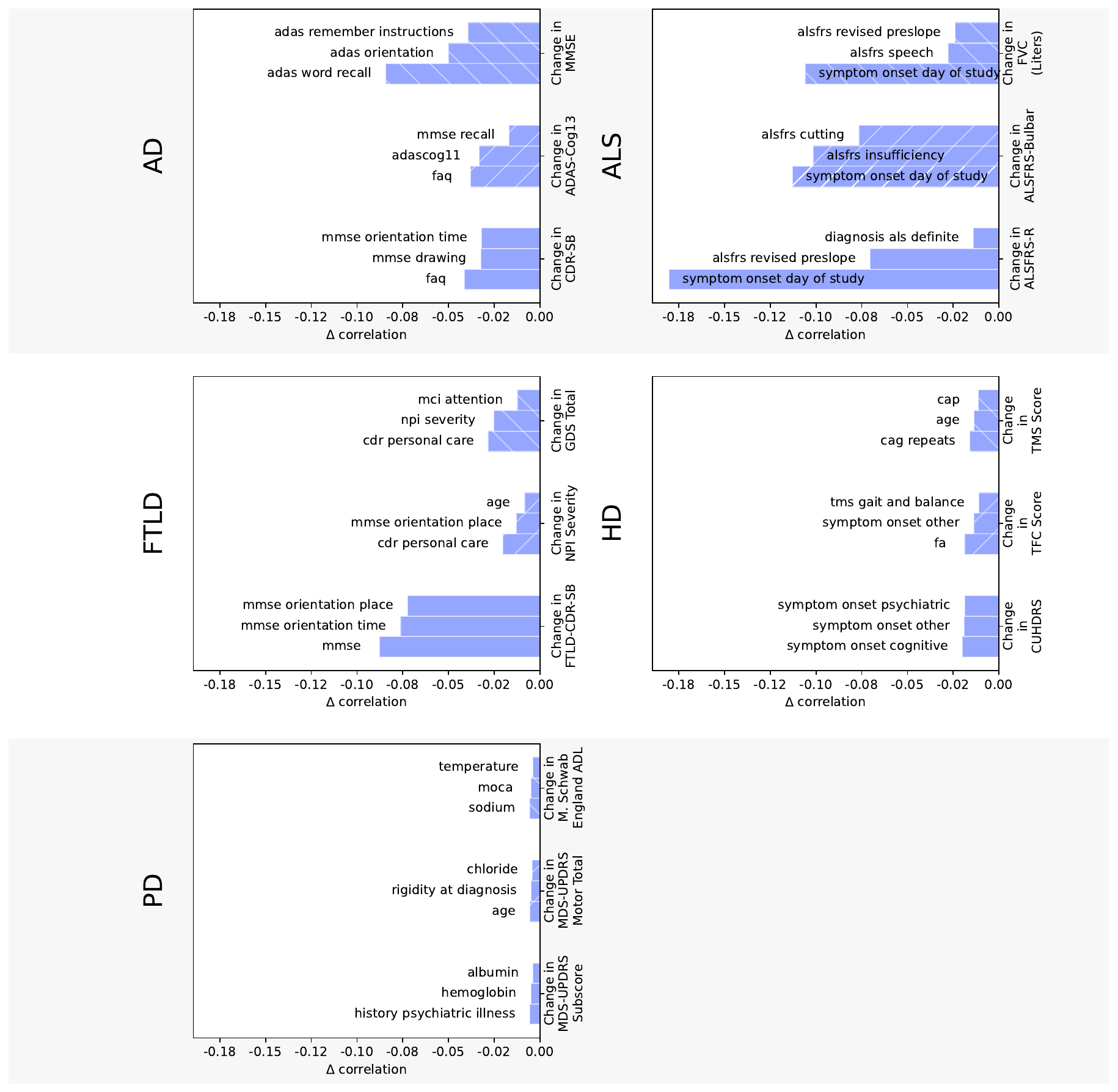}
    \caption{Input Sensitivity for the Neurodegeneration therapeutic Area. For each indication, three important outcomes are selected and indicate in the y-axis label. Bars indicate how much certain features indicated in the text on the plot reduce the correlation of the feature's prediction when evaluated on the entire population of held out patients.}
    \label{fig:inpsen_neuro}
\end{figure}

\begin{figure}
    \centering
    \includegraphics[width=14cm, trim = 3.75cm 0cm 2cm 0cm, clip]{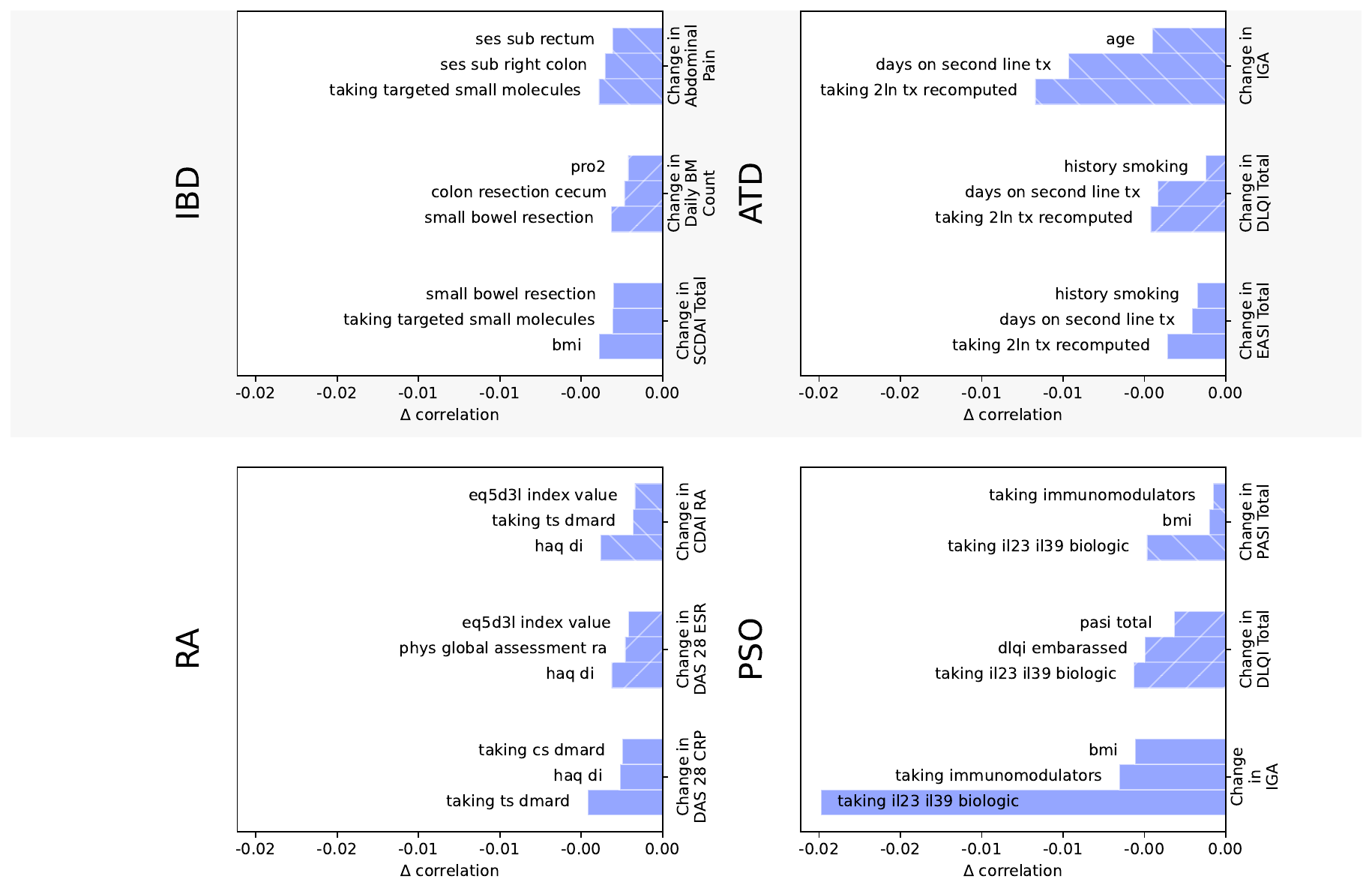}
    \caption{Input Sensitivity for the Immunology \& Inflammation therapeutic Area. For each indication, three important outcomes are selected and indicate in the y-axis label. Bars indicate how much certain features indicated in the text on the plot reduce the correlation of the feature's prediction when evaluated on the entire population of held out patients.}
    \label{fig:inpsen_ini}
\end{figure}

\begin{figure}
    \centering
    \includegraphics[width=14cm, trim = 3.75cm 0cm 2cm 0cm, clip]{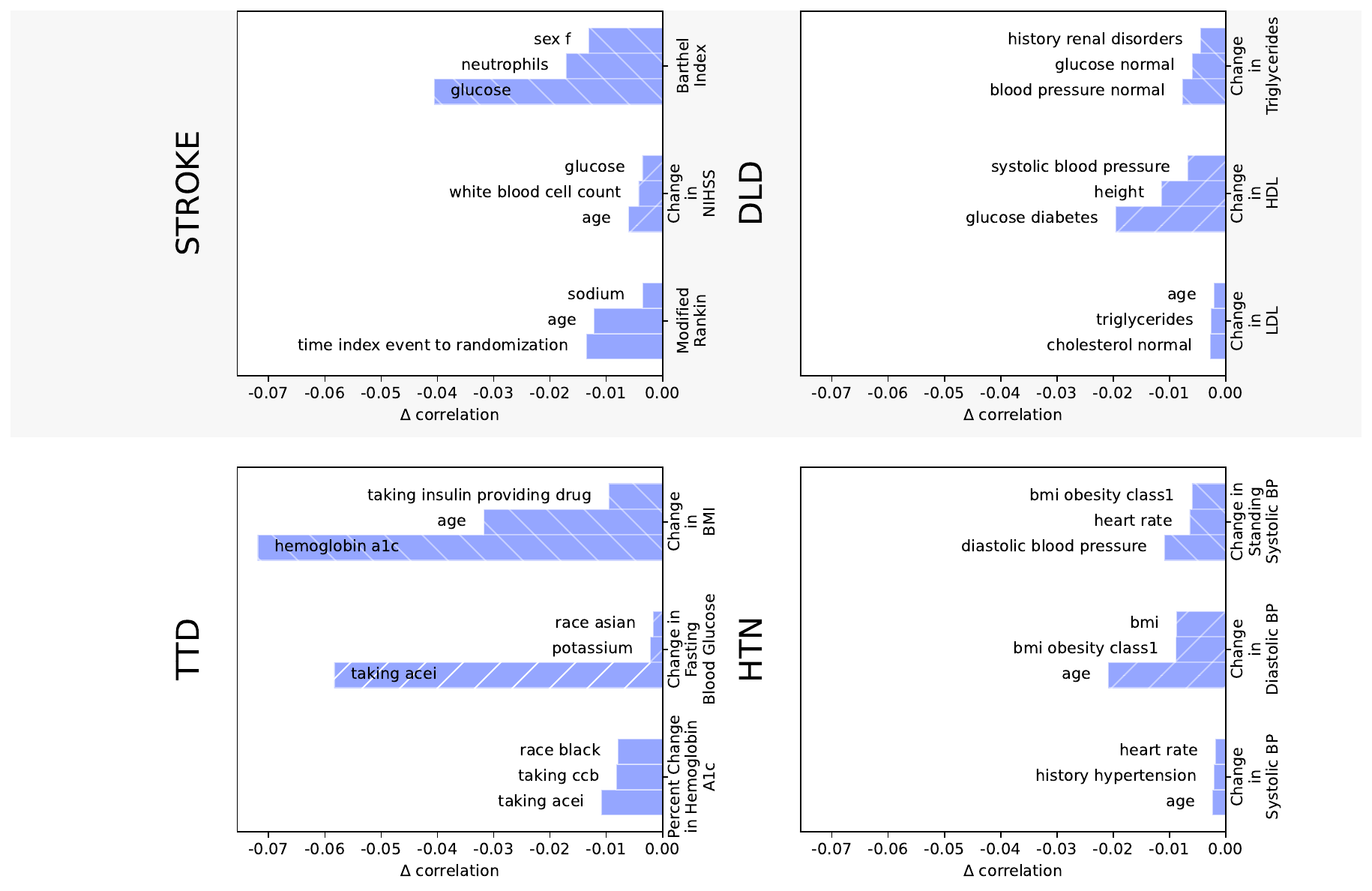}
    \caption{Input Sensitivity for the General Medicine therapeutic Area. For each indication, three important outcomes are selected and indicate in the y-axis label. Bars indicate how much certain features indicated in the text on the plot reduce the correlation of the feature's prediction when evaluated on the entire population of held out patients.}
    \label{fig:inpsen_gm}
\end{figure}

When using machine learning models to make important decisions, it is useful to develop an understanding about how the models work and how sensitive their outputs may be to changes in their inputs. We use a couple different methods to probe our models. The first we call input sensitivity (see \Fig{inpsen_neuro}, \Fig{inpsen_ini}, and \Fig{inpsen_gm}), which measures how much a given performance metric, for instance Pearson correlation between data and mean twins over a certain cohort, changes when a given feature is masked from the patient's baseline data. Missing data is a common problem in many healthcare applications--that is, some assessments may not be available as model inputs because they are too expensive, time-consuming, or invasive to collect on every patient--so how robust model performance is to missing inputs is an important characteristic. In certain indications like PD, we see that input sensitivity is low relative to other indications in the Neurodegeneration therapeutic area. This can arise when the feature being probed is not strongly correlated to other features and instead depends primarily on its own baseline value. For the Immunology \& Inflammation therapeutic area, we see that generally all the probed outcomes have dominant dependence on their baseline value and thus the change in correlation to masking other variables is small. The General Medicine therapeutic area falls in between Neurodegeneration and Immunology \& Inflammation in this regard.

\begin{figure}
    \centering
    \includegraphics[width=14cm, trim = 0.5cm 0cm 2cm 0cm, clip]{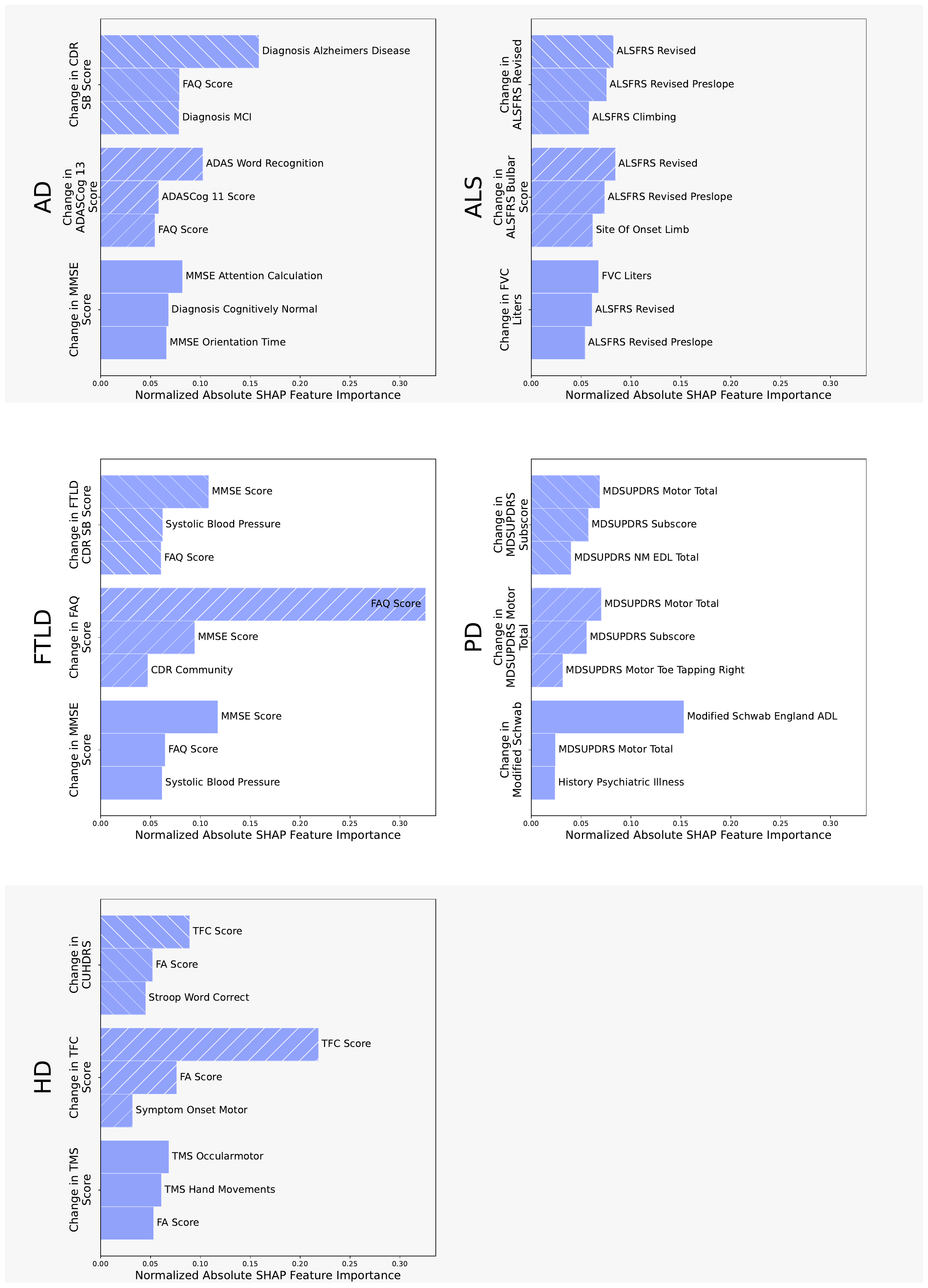}
    \caption{SHAP values for the Neurodegeneration therapeutic Area. For each patient, SHAP values for each baseline feature are first taken as absolute values, then normalized so that their sum equals one. This process is repeated for each of three important outcomes for each indication indicated in the y-axis label. Finally, these normalized SHAP values are averaged across all patients to derive the average contribution of each feature per outcome. }
    \label{fig:shap_neuro}
\end{figure}

\begin{figure}
    \centering
    \includegraphics[width=14cm, trim = 0.5cm 0cm 2cm 0cm, clip]{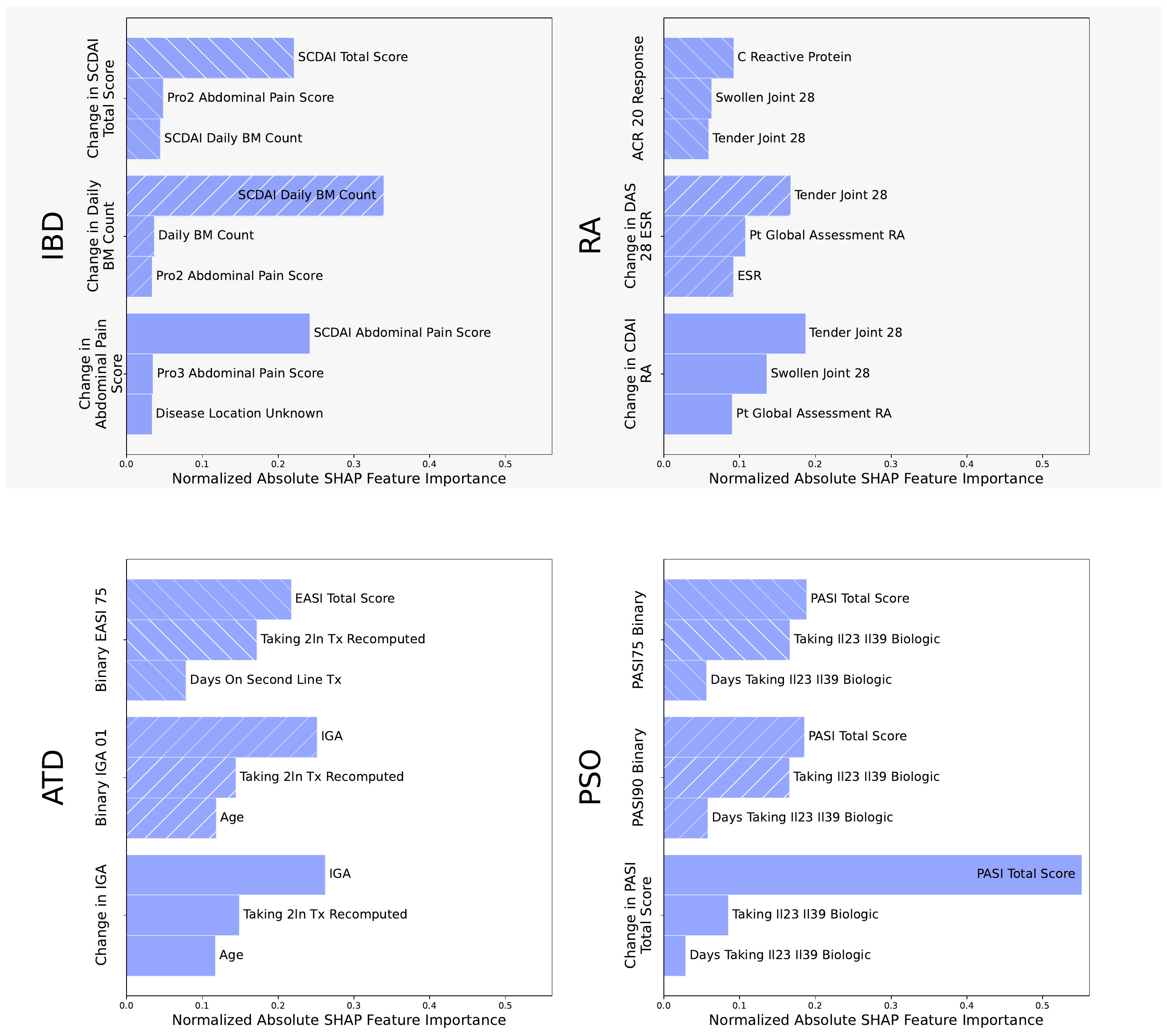}
    \caption{SHAP values for the Immunology \& Inflammation therapeutic Area. For each patient, SHAP values for each baseline feature are first taken as absolute values, then normalized so that their sum equals one. This process is repeated for each of three important outcomes for each indication indicated in the y-axis label. Finally, these normalized SHAP values are averaged across all patients to derive the average contribution of each feature per outcome. }
    \label{fig:shap_ini}
\end{figure}

\begin{figure}
    \centering
    \includegraphics[width=14cm, trim = 0.5cm 0cm 2cm 0cm, clip]{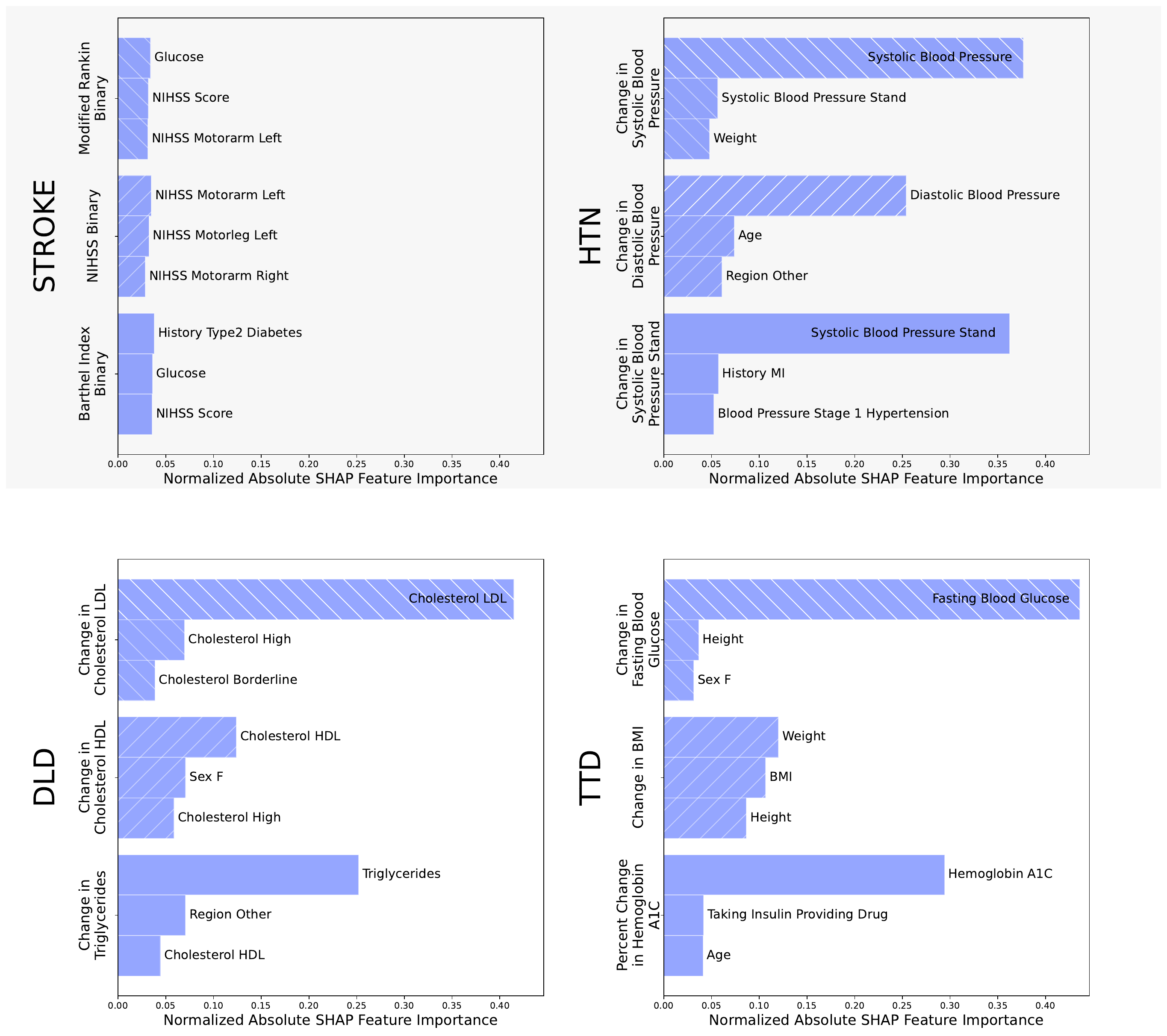}
    \caption{SHAP values for the General Medicine therapeutic Area. For each patient, SHAP values for each baseline feature are first taken as absolute values, then normalized so that their sum equals one. This process is repeated for each of three important outcomes for each indication indicated in the y-axis label. Finally, these normalized SHAP values are averaged across all patients to derive the average contribution of each feature per outcome.}
    \label{fig:shap_gm}
\end{figure}

In addition to input sensitivity, we use SHAP (SHapley Additive exPlanations) to gain deeper insights into our models decisions. SHAP provides a detailed view of how each feature contributes to individual predictions, regardless of how well that prediction matches to the data. This distinction is crucial as SHAP reveals the influence of each feature within the context of a specific prediction, allowing us to understand the "why" behind a model's output in a way that input sensitivity does not. SHAP thus offers a complementary perspective, focusing on prediction contribution rather than performance sensitivity. SHAP is also able to probe the effect of the baseline value on the prediction, whereas Input Sensitivity can only look at the effect of other outcomes on a given outcome. In \Fig{shap_neuro}, \Fig{shap_ini}, and \Fig{shap_gm} we report SHAP values for key outcomes across indications for each of our DTGs. We generally find that SHAP values agree with domain knowledge of a given indication.

\subsection{TTE}

\begin{figure}
    \centering
    \includegraphics[width=10cm, clip]{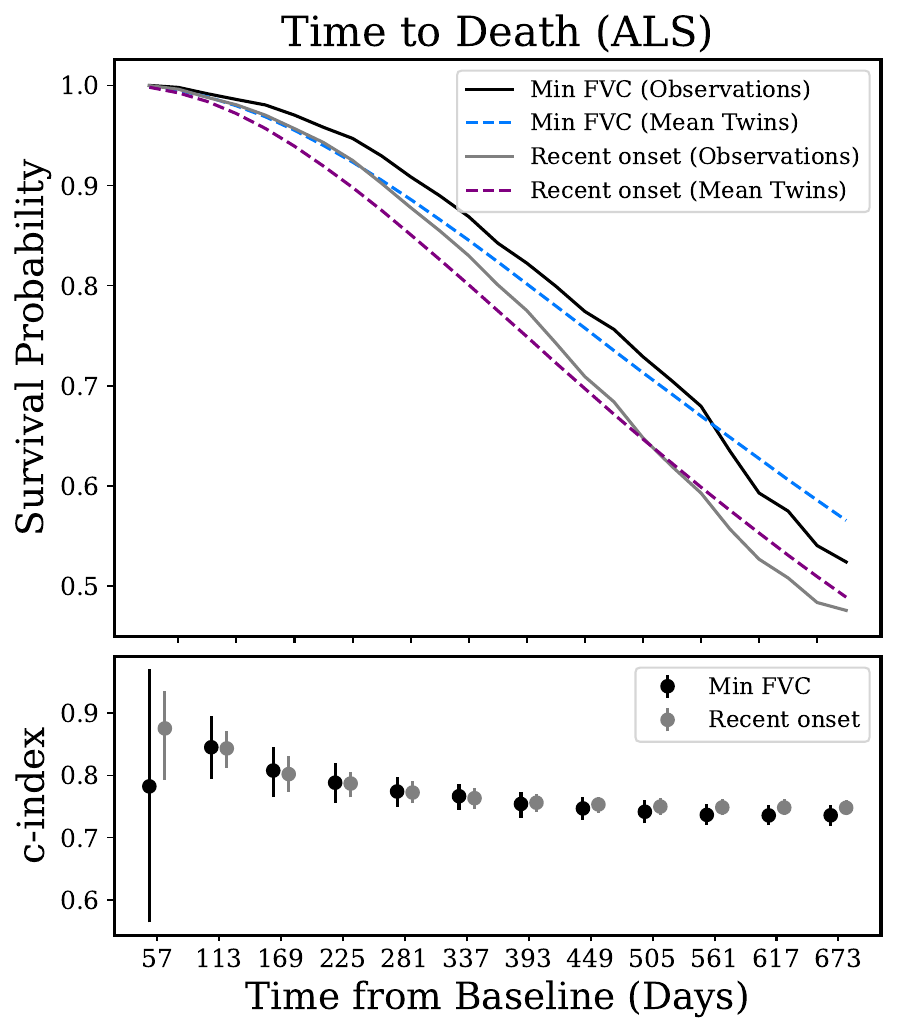}
    \caption{Evaluation of the TTE model for ALS time to death outcome. In the {\bf top panel} we 
    compare the observed and predicted survival curves for the two top cohorts defined in \Tab{cohort-def-neuro}. The {\bf bottom panel} reports concordance index as a function of time. Concordance index at a given time assumes that all observed events after that time are censored. Error bars indicate 95\% confidence intervals.}
    \label{fig:survival}
\end{figure}

DTGs can also model survival outcomes, and in
\Fig{survival}, we report a comparison of the observed and predicted survival
curves for the time-to-death outcome in ALS for two clinically relevant cohorts. We also report concordance index as a measure of goodness-of-fit beyond population-level survival curves.

\section{Future Directions}
\label{sec:future}

In this technical report, we have shown how Digital Twin Generators (DTGs) in 13 separate indications can be trained to forecast patient outcomes in a specified disease area, all with a common neural network architecture called the Neural Boltzmann Machine (NBM). There are many potential avenues for the improvement of these methodologies, such as increasing the size of the datasets, incorporating synthetic data or data augmentation techniques during training, or exploring alternative neural network architectures within the NBM. 

Although we've shown that it's possible to train these DTGs using the same general architecture and minimal hyperparameter tuning, there is substantial effort required per indication to develop each disease-specific DTG, particularly in the curation of the training datasets. In principle, there is a great deal of information that is shared across indications that may be useful to inform models of disease progression within each indication through transfer learning.  To fully capture that shared information, a significant advancement in AI applications for healthcare would be the development of  a \emph{universal} DTG. This is a single model able to serve as a general predictor of disease progression for a wide range of indications. We regard such a model as a foundational model for digital twins, as many other applications can be built using it as a computational substrate.  Like the development of foundational models in other applications, models of this type for digital twins may evolve over time to accept and make use of a broader set of input data such as electronic health records (EHR), genomic data, wearables, imaging, and more. 

While the DTGs described in this report are trained to generate clinical records in a single indication, they all require and only make use of the same type of tabular longitudinal data, and each is trained only with the data from that indication.  This is limiting in terms of the scope of 
any individual DTG since, for example, the Frontotemporal Dementia DTG cannot learn from the more abundant data from other neurodegenerative diseases.  Furthermore, the architecture used here, at least as implemented, does not directly permit models to be trained with other data sources, such as summary data from populations, text-based domain knowledge, or even differently formatted tabular data, all of which are useful sources of information for modeling disease. We note, however, that the probabilistic continuous time modeling core that the NBM represents can generalize over multiple input modalities -- one need only change the neural networks that parameterize the NBM to accept new data formats. We expect that the path to creating a universal DTG follows the same recipe as used in the development of other foundation models: mapping all potential inputs into a shared latent representation, modeling the stochastic dynamics within that latent representation, and then mapping back to the visible space. 

The broader goal of using AI tools to improve outcomes in medicine requires models that can serve many applications, such as comparative drug effectiveness, health outcomes research, and clinical decision support.  Several challenges emerge for building universal models that can
forecast patient outcomes across multiple indications. To truly achieve a universal disease progression model, AI systems must be scalable and easily adaptable to different healthcare infrastructures and population demographics. This includes the ability to update models with new
data continuously and adapt to emerging diseases and changing epidemiological
patterns. In the realm of clinical trials, changes to the standard-of-care or
novel clinical assessments must be able to be seamlessly integrated into these models to ensure that new treatments are assessed with up-to-date information.  Perhaps somewhat paradoxically, these kinds of updates are much simpler with foundation models, as they are able to intake and use data more flexibly for training, can be fine-tuned for new applications with relatively small datasets, and require significantly less maintenance. 

To address these challenges, we're working towards developing universal DTGs with capabilities for zero-shot conditioning and prediction by leveraging broad, multi-indication datasets that include as wide of a variety of data sources as possible while still maintaining, or improving on, the quality of the resulting model.

\section{Conclusion}
\label{sec:conclusions}

In this report, we have described a neural network architecture used to create Digital Twin Generators (DTGs) in 13 different disease indications across 3 therapeutic areas, as well as properties of the methods and data used to build and evaluate the DTGs. This work showcases the ability of modern machine learning architectures for learning to create digital twins of individual patients that can forecast their future health in detail, paving the way for applications in clinical research, comparative effective studies, and personalized medicine. In addition, this work contributes to the broader machine learning literature by (i) presenting a new application of a recently introduced class energy-based models called Neural Boltzmann Machines and (ii) describing a new approach to probabilistic modeling of multivariate continuous-time dynamical systems. 

Pursuing follow-up work in multi-indication models brings us closer to realizing an AI-powered universal model that not only predicts diseases across multiple indications, but also enhances the accessibility and quality of healthcare worldwide. By using AI to improve upon the status quo, we can
innovate a new "gold standard" for assessing the safety and efficacy of new
treatments, paving the way for AI-enabled personalized medicine to become a
global reality.

\section{Contributors}

Nameyeh Alam, 
Jake Basilico, 
Daniele Bertolini, 
Satish Casie Chetty, 
Heather D'Angelo,
Ryan Douglas, 
Charles K. Fisher, 
Franklin Fuller, 
Melissa Gomes,
Rishabh Gupta, 
Alex Lang,
Anton Loukianov, 
Rachel Mak-McCully, 
Cary Murray,
Hanalei Pham, 
Susanna Qiao, 
Elena Ryapolova-Webb, 
Aaron Smith, 
Dimitri Theoharatos, 
Anil Tolwani, 
Eric W. Tramel, 
Anna Vidovszky, 
Judy Viduya, 
and 
Jonathan R. Walsh.

\section{Data Acknowledgements}
The data sources used to train and test the models described in this report are acknowledged here: \href{https://www.unlearn.ai/data-acknowledgements}{https://www.unlearn.ai/data-acknowledgements}

\newpage

\bibliographystyle{plain} 
\bibliography{references} 

\newpage


\newpage
\section{Appendix}

Some supplemental tables with indication specific definitions of outcomes and cohorts (sub-populations of a dataset) are provided here, broken down by therapeutic area. We also report example digital
twins for individual patients in each indication, showing the predicted mean and standard deviation through time for all variables.

\begin{table}
    \centering
    \begin{tabularx}{5.5in}{XX}\hline
        \toprule
        \multicolumn{2}{>{\hsize=\dimexpr2\hsize+2\tabcolsep+\arrayrulewidth\relax}X}
        {\textbf{Disease Indication}: AD} \\
        \midrule
        \textbf{Outcome}  &  \textbf{Description}\\
        CDR-SB change from baseline  & Clinical Dementia Rating Scale Sum of Boxes. It assesses stages of dementia.\\
        ADASCog-13 change from baseline  & Alzheimer's Disease Assessment Scale cognitive subscale. It measures cognitive function.\\
        MMSE change from baseline  & Mini Mental State Examination. It measures cognitive function.\\
        \midrule
        \multicolumn{2}{>{\hsize=\dimexpr2\hsize+2\tabcolsep+\arrayrulewidth\relax}X}{\textbf{Disease Indication}: ALS} \\
        \midrule
        \textbf{Outcome}  &  \textbf{Description}\\
        ALSFRS Revised change from baseline & Revised Amyotrophic Lateral Sclerosis Functional Rating Scale. It measures severity of ALS.\\
        ALSFRS Revised Bulbar Subscore change from baseline & Bulbar Subscore of ALSFRS Revised. It measures impairments to speech, salivation, and swallowing.\\
        FVC change from baseline & Forced Vital Capacity. It measures breathing function.\\
        \midrule
        \multicolumn{2}{>{\hsize=\dimexpr2\hsize+2\tabcolsep+\arrayrulewidth\relax}X}{\textbf{Disease Indication}: FTLD} \\
        \midrule
        \textbf{Outcome}  &  \textbf{Description}\\
        FTLD CDR-SB change from baseline & FTLD version of the Clinical Dementia Rating Scale Sum of Boxes. It measures stages of dementia.\\
        FAQ change from baseline & Functional Activities Questionnaire which measures functional problems in activities of daily living.\\
        MMSE change from baseline  & Mini Mental State Examination. It assesses cognitive function.\\
        \midrule
        \multicolumn{2}{>{\hsize=\dimexpr2\hsize+2\tabcolsep+\arrayrulewidth\relax}X}{\textbf{Disease Indication}: PD} \\
        \midrule
        \textbf{Outcome}  &  \textbf{Description}\\
        MDSUPDRS Subscore change from baseline & Movement Disorder Society - Unified Parkinson’s Disease Rating Scale. It evaluates motor and non-motor symptoms of Parkinson’s disease.\\
        MDSUPDRS Motor Subscore change from baseline & The motor component of the MDSUPDRS.\\
        Modified Schwab and England change from baseline & It measures abilities in a variety of activities of daily living.\\
        \midrule
        \multicolumn{2}{>{\hsize=\dimexpr2\hsize+2\tabcolsep+\arrayrulewidth\relax}X}{\textbf{Disease Indication}: HD} \\
        \midrule
        \textbf{Outcome}  &  \textbf{Description}\\
        cUHDRS change from baseline & Composite Unified Huntington Disease Rating Scale. It measures motor, cognitive, and functional decline.\\
        TFC change from baseline & Total Functional Capacity. It measures functional decline.\\
        TMS change from baseline & Total Motor Score. It measures motor decline.\\
        \bottomrule
    \end{tabularx}
    \caption{Brief description of outcomes for indications in the Neurodegenerative Therapeutic Area.}
    \label{tab:outcomes-neuro}
\end{table}

\begin{table}
    \centering
    \begin{tabularx}{5in}{XX}\hline
        \toprule
        \multicolumn{2}{>{\hsize=\dimexpr2\hsize+2\tabcolsep+\arrayrulewidth\relax}X}
        {\textbf{Disease Indication}: CD} \\
        \midrule
        \textbf{Outcome}  &  \textbf{Description}\\
        sCDAI change from baseline & Short Crohn's Disease Activity Index. It measures current severity of the disease. \\
        Daily BM Count change from baseline & Count of bowel movements.\\
        Abdominal Pain Score change from baseline & It measures intensity of abdominal pain.\\
        \midrule
        \multicolumn{2}{>{\hsize=\dimexpr2\hsize+2\tabcolsep+\arrayrulewidth\relax}X}{\textbf{Disease Indication}: RA} \\
        \midrule
        \textbf{Outcome}  &  \textbf{Description}\\
        Binary ACR 20 Response & A composite measure of whether the participant improves by at least 20\% in both swollen and tender joint counts and at least three of five additional disease criteria.\\
        DAS-28 ESR change from baseline & Disease Activity Score-28 for Rheumatoid Arthritis with ESR. It measures severity of rheumatoid arthritis using clinical and laboratory data, including Erythrocyte Sedimentation Rate.\\
        CDAI-RA change from baseline & Clinical Disease Activity Index for Rheumatoid Arthritis. It measures severity of rheumatoid arthritis using clinical data only.\\
        \midrule
        \multicolumn{2}{>{\hsize=\dimexpr2\hsize+2\tabcolsep+\arrayrulewidth\relax}X}{\textbf{Disease Indication}: ATD} \\
        \midrule
        \textbf{Outcome}  &  \textbf{Description}\\
        Binary EASI-75 & Binarized version of Eczema Area and Severity Index corresponding to at least 75\% reduction from baseline. It measures extent and severity of atopic dermatitis.\\
        Binary IGA-01 & Binarized version of investigator global assessment. \\
        IGA change from baseline & Investigator global assessment. It measures severity of atopic dermatitis.\\
        \midrule
        \multicolumn{2}{>{\hsize=\dimexpr2\hsize+2\tabcolsep+\arrayrulewidth\relax}X}{\textbf{Disease Indication}: PSO} \\
        \midrule
        \textbf{Outcome}  &  \textbf{Description}\\
        Binary PASI-75 & Binarized version of Psoriasis Area and Severity Index. It represents an improvement of at least 75\% from baseline.\\
        Binary PASI-90 & Binarized version of Psoriasis Area and Severity Index. It represents an improvement of at least 90\% from baseline.\\
        PASI change from baseline & Psoriasis Area and Severity Index. It measures severity of psoriasis.\\
        \bottomrule
    \end{tabularx}
    \caption{Brief description of outcomes for indications in the Immunology \& Inflammation Therapeutic Area.}
    \label{tab:outcomes-ini}
\end{table}

\begin{table}
    \centering
    \begin{tabularx}{5in}{XX}\hline
        \toprule
        \multicolumn{2}{>{\hsize=\dimexpr2\hsize+2\tabcolsep+\arrayrulewidth\relax}X}
        {\textbf{Disease Indication}: STR} \\
        \midrule
        Binary Modified Rankin & Binarized version of Modified Rankin Scale. It measures degree of disability in the activities of daily living as a result of stroke.\\
        Binary NIHSS & Binarized version of the National Institutes of Health Stroke Scale. It measures stroke severity.\\
        Binary Barthel Index & Binarized version of Barthel Index. It measures degree of disability in the activities of daily living as a result of stroke.\\
        \midrule
        \multicolumn{2}{>{\hsize=\dimexpr2\hsize+2\tabcolsep+\arrayrulewidth\relax}X}{\textbf{Disease Indication}: HTN} \\
        \midrule
        \textbf{Outcome}  &  \textbf{Description}\\
        Systolic blood pressure change from baseline & The systolic blood pressure.\\
        Diastolic blood pressure change from baseline & The diastolic blood pressure.\\
        Standing systolic blood pressure change from baseline & The systolic blood pressure measures while standing.\\
        \midrule
        \multicolumn{2}{>{\hsize=\dimexpr2\hsize+2\tabcolsep+\arrayrulewidth\relax}X}{\textbf{Disease Indication}: DLD} \\
        \midrule
        \textbf{Outcome}  &  \textbf{Description}\\
        Cholesterol LDL change from baseline & Fasting low-density lipoprotein cholesterol.\\
        Cholesterol HDL change from baseline & Fasting high-density lipoprotein cholesterol.\\
        Triglycerides change from baseline & Fasting triglycerides.\\
        \midrule
        \multicolumn{2}{>{\hsize=\dimexpr2\hsize+2\tabcolsep+\arrayrulewidth\relax}X}{\textbf{Disease Indication}: TTD} \\
        \midrule
        \textbf{Outcome}  &  \textbf{Description}\\
        Hemoglobin A1c percent change from baseline & It measures average blood sugar levels over the past 3 months.\\
        Fasting blood glucose change from baseline & It measures blood glucose after at least 8 hours without eating.\\
        BMI change from baseline & Body Mass Index.\\
        \bottomrule
    \end{tabularx}
    \caption{Brief description of outcomes for indications in the General Medicine Therapeutic Area.}
    \label{tab:outcomes-gm}
\end{table}

\begin{table}
    \centering
    \begin{tabularx}{5in}{XX}\hline
        \toprule
        \multicolumn{2}{>{\hsize=\dimexpr2\hsize+2\tabcolsep+\arrayrulewidth\relax}X}{\textbf{Disease Indication}: AD} \\
        \midrule
        \textbf{Cohort}  & \textbf{Description} \\
        \textsuperscript{$\star$} Mild Cognitive Impairment (MCI) and early Alzheimer's Disease (AD) & \tabitem 50 $\leq$ age $\leq$ 90 \\
        & \tabitem One of the following: \\
        & \;\;\, \tabitemii 20 $\leq$ MMSE $\leq$ 28 \\
        & \;\;\, \tabitemii OR 0 < CDR-GS $\leq$ 1 \\
        Mild to moderate Alzheimer's Disease (AD) & \tabitem 50 $\leq$ age $\leq$ 90 \\
        & \tabitem 13 $\leq$ MMSE $\leq$ 26 \\
        Full Population & All participants in the dataset \\
        \midrule
        \multicolumn{2}{>{\hsize=\dimexpr2\hsize+2\tabcolsep+\arrayrulewidth\relax}X}{\textbf{Disease Indication}: ALS} \\
        \midrule
        \textbf{Cohort}  & \textbf{Description} \\
        \textsuperscript{$\star$} Recent Onset of Symptoms & ALS symptoms recently (within 24 months) developed since baseline measurement \\
        Minimum Vital Capacity & Cohort has FVC $\geq$ 60\% of normal value \\
        Full Population & All participants in the dataset \\
        \midrule
        \multicolumn{2}{>{\hsize=\dimexpr2\hsize+2\tabcolsep+\arrayrulewidth\relax}X}{\textbf{Disease Indication}: FTD} \\
        \midrule
        \textbf{Cohort}  & \textbf{Description} \\
        \textsuperscript{$\star$} Has behavioral variant of Frontotemporal Dementia & As judged in primary diagnosis of patient \\
        Has known FTD Mutation & patient has a mutation in either MAPT, PGRN, or C9orf72 genes \\
        Full Population & All participants in the dataset \\
        \midrule
        \multicolumn{2}{>{\hsize=\dimexpr2\hsize+2\tabcolsep+\arrayrulewidth\relax}X}{\textbf{Disease Indication}: HD} \\
        \midrule
        \textbf{Cohort}  & \textbf{Description} \\
        \textsuperscript{$\star$} Manifest & Has HTT expansion (CAG repeats) and clinical symptoms of HD \\
        Full Population & All participants in the dataset \\
        \midrule
        \multicolumn{2}{>{\hsize=\dimexpr2\hsize+2\tabcolsep+\arrayrulewidth\relax}X}{\textbf{Disease Indication}: PD} \\
        \midrule
        \textbf{Cohort}  & \textbf{Description} \\
        \textsuperscript{$\star$} Has Early PD & Patient is off symptom-modifying therapies and is early stage (Hoehn and Yahr scale <= 2) \\
        Off SMT & Patient is off symptom-modifying therapies at baseline \\
        On SMT & Patient is off symptom-modifying therapies at baseline \\
        \bottomrule
    \end{tabularx}
    \caption{Cohort Definitions for Neurodegeneration Therapeutic Area. The cohort indicated with a \textsuperscript{$\star$} indicates the principal cohort for the  indication. Unless specified otherwise, all plots here report metrics on the principal cohort.}
    \label{tab:cohort-def-neuro}
\end{table}

\begin{table}
    \centering
    \begin{tabularx}{5in}{XX}\hline
        \toprule
        \multicolumn{2}{>{\hsize=\dimexpr2\hsize+2\tabcolsep+\arrayrulewidth\relax}X}{\textbf{Disease Indication}: CD} \\
        \midrule
        \textbf{Cohort}  & \textbf{Description} \\
        \textsuperscript{$\star$} Moderate to Severe Crohn's Disease & sCDAI Score $\ge$ 225 \\
        Diagnosed with Crohn's Disease & Clinical Diagnosis of Crohn's Disease \\
        \midrule
        \multicolumn{2}{>{\hsize=\dimexpr2\hsize+2\tabcolsep+\arrayrulewidth\relax}X}{\textbf{Disease Indication}: RA} \\
        \midrule
        \textbf{Cohort}  & \textbf{Description} \\
        \textsuperscript{$\star$} Moderate to Severe & patients with moderate to severe disease activity based on DAS28-CRP~(>~2.9) or DAS28-ESR~(>~3.2) or SDAI~(>~20) or CDAI~(>~10) \\
        Moderate to Severe: DMARD Resistant & Moderate to severe patients who have been treated with a Disease Modifying Anti-Rheumatic Drug (DMARD) at least 3 months prior to study start \\
        Moderate to Severe: on Stable Methotrexate & Moderate to severe patients who have been on Methotrexate for at least 3 months \\
        \midrule
        \multicolumn{2}{>{\hsize=\dimexpr2\hsize+2\tabcolsep+\arrayrulewidth\relax}X}{\textbf{Disease Indication}: ATD} \\
        \midrule
        \textbf{Cohort}  & \textbf{Description} \\
        \textsuperscript{$\star$} Moderate to Severe AtD & Patients with: \\
            & \tabitem EASI Total Score $\geq$ 16 \\
            & \tabitem IGA $\geq$ 3 \\
            & \tabitem BSA $\geq$ 10\% \\
        Mild to Moderate AtD & Patients with: \\
            & \tabitem 1.1 $\leq$ EASI Total Score $\leq$ 21 \\
            & \tabitem 2 $\leq$ IGA $\leq$ 3 \\
            & \tabitem 3\% $\leq$ BSA $\leq$ 20\% \\
        Novel Disease-Modifying Antirheumatic Drug (DMARD) use & Patients with: \\
            & \tabitem Initiate a second line treatment (biologics, DMARDs, or immunomodulators) on baseline day. \\
            & \tabitem At least 90 days since ending prior second line treatment \\
        \midrule
        \multicolumn{2}{>{\hsize=\dimexpr2\hsize+2\tabcolsep+\arrayrulewidth\relax}X}{\textbf{Disease Indication}: PSO} \\
        \midrule
        \textbf{Cohort}  & \textbf{Description} \\
        \textsuperscript{$\star$} Full Population & All participants in the dataset \\
        Moderate to Severe Plaque Psoriasis & Patients with: \\
        & \tabitem PASI Total Score $\geq$ 12, \\
        & \tabitem IGA $\geq$ 3, \\
        & \tabitem BSA $\geq$ 10\%, \\
        & \tabitem no diagnosis of Psoriatic Arthritis. \\
        \bottomrule
    \end{tabularx}
    \caption{Cohort Definitions for Immunology \& Inflammation Therapeutic Area. The cohort indicated with a \textsuperscript{$\star$} indicates the principal cohort for the  indication. Unless specified otherwise, all plots here report metrics on the principal cohort.}
    \label{tab:cohort-def-ini}
\end{table}

\begin{table}
    \centering
    \begin{tabularx}{5.5in}{XX}\hline
        \toprule
        \multicolumn{2}{>{\hsize=\dimexpr2\hsize+2\tabcolsep+\arrayrulewidth\relax}X}{\textbf{Disease Indication}: STR} \\
        \midrule
        \textbf{Cohort}  & \textbf{Description} \\
        \textsuperscript{$\star$} Full Population & All participants in the dataset \\
        \midrule
        \multicolumn{2}{>{\hsize=\dimexpr2\hsize+2\tabcolsep+\arrayrulewidth\relax}X}{\textbf{Disease Indication}: DLD} \\
        \midrule
        \textbf{Cohort}  & \textbf{Description} \\
        \textsuperscript{$\star$} Hyperlipidemia (HLD) & Participants on statins with \\
           & \tabitem History of dyslipidemia OR\\
           & \tabitem 100 mg/dL $\leq$ LDL-C < 190 mg/dL, triglycerides $\leq$ 400 mg/dL, and age over 18 years \\
           & Excludes participants with \\
           & \tabitem Uncontrolled hypertension (blood pressure $\geq$ 140/90 mmHg) OR\\
           & \tabitem History of coronary heart disease, heart failure, myocardial  infarction, or stroke \\
        Atherogenic Dyslipidemia & Participants on statins with vascular disease \\
            & and atherogenic dyslipidemia defined as \\
            & \tabitem LDL-C $\leq$ 160 mg/dL \\
            & \tabitem 150 mg/dL $\leq$ triglycerides $\leq$ 400 mg/dL \\
            & \tabitem HDL-C $\leq$ 40 mg/dL for men or $\leq$ 50 mg/dL for women \\
        \midrule
        \multicolumn{2}{>{\hsize=\dimexpr2\hsize+2\tabcolsep+\arrayrulewidth\relax}X}{\textbf{Disease Indication}: HTN} \\
        \midrule
        \textbf{Cohort}  & \textbf{Description} \\
        \textsuperscript{$\star$} On Hypertension Medication & Patients taking at least one antihypertensive medication \\
        Population with Hypertension & Participants with \\
            & \tabitem a history of hypertension or blood pressure $\geq$ 130/80 \\
            & \tabitem are older than 18 \\
            & \tabitem not had history of: stroke, unstable angia, myocardial infarction, or heart failure \\
        \midrule
        \multicolumn{2}{>{\hsize=\dimexpr2\hsize+2\tabcolsep+\arrayrulewidth\relax}X}{\textbf{Disease Indication}: TTD} \\
        \midrule
        \textbf{Cohort}  & \textbf{Description} \\
        \textsuperscript{$\star$} Comorbid with HLD or HTN & Participants with \\
            & \tabitem History of Type-2 Diabetes Mellitus or HbA1c $\geq$ 7.0\% and either \\
            & \tabitem History of hyperlipidemia or LDL-C $\geq$ 130 mg/dL, OR \\
            & \tabitem History of hypertension or systolic blood pressure $\geq$ 130 mmHg \\
        Comorbid Coronary Artery Disease & Participants with \\
            & \tabitem History of Type-2 Diabetes Mellitus or HbA1c $\geq$ 7.0\% \\
            & \tabitem History of coronary artery disease: either ischemic heart disease or acute events such as myocardial infarction \\
        \bottomrule
    \end{tabularx}
    \caption{Cohort Definitions for the General Medicine Therapeuic Area. The cohort indicated with a \textsuperscript{$\star$} indicates the principal cohort for the  indication. Unless specified otherwise, all plots here report metrics on the principal cohort.}
    \label{tab:cohort-def-gm}
\end{table}

\end{document}